\documentclass[11pt]{article}
\usepackage{amsmath}
\usepackage{amssymb}
\usepackage{arydshln}
\usepackage{authblk}
\usepackage{booktabs}
\usepackage[margin=1in]{geometry}
\usepackage{graphicx}
\usepackage{multirow}
\usepackage[sort,numbers]{natbib}
\usepackage{xcolor}
\usepackage[colorlinks=true,citecolor=blue]{hyperref}

\graphicspath{{figs/}}

\title{Fourier-DeepONet: Fourier-enhanced deep operator networks for full waveform inversion with improved accuracy, generalizability, and robustness}
\author[1,2]{Min Zhu}
\author[3]{Shihang Feng}
\author[2]{Youzuo Lin}
\author[1,*]{Lu Lu}
\affil[1]{Department of Statistics and Data Science, Yale University, New Haven, CT 06511, USA}
\affil[2]{Earth and Environmental Sciences Division, Los Alamos National Laboratory, Los Alamos, NM 87544, USA}
\affil[3]{Theoretical Division, Los Alamos National Laboratory, Los Alamos, NM 87544, USA}
\affil[*]{Corresponding author. Email: lulu1@seas.upenn.edu}

\date{}

\begin{document}

\maketitle

\begin{abstract}
Full waveform inversion (FWI) infers the subsurface structure information from seismic waveform data by solving a non-convex optimization problem. Data-driven FWI has been increasingly studied with various neural network architectures to improve accuracy and computational efficiency. Nevertheless, the applicability of pre-trained neural networks is severely restricted by potential discrepancies between the source function used in the field survey and the one utilized during training. Here, we develop a Fourier-enhanced deep operator network (Fourier-DeepONet) for FWI with the generalization of seismic sources, including the frequencies and locations of sources. Specifically, we employ the Fourier neural operator as the decoder of DeepONet, and we utilize source parameters as one input of Fourier-DeepONet, facilitating the resolution of FWI with variable sources. To test Fourier-DeepONet, we develop three new and realistic FWI benchmark datasets (FWI-F, FWI-L, and FWI-FL) with varying source frequencies, locations, or both. Our experiments demonstrate that compared with existing data-driven FWI methods, Fourier-DeepONet obtains more accurate predictions of subsurface structures in a wide range of source parameters. Moreover, the proposed Fourier-DeepONet exhibits superior robustness when handling data with Gaussian noise or missing traces and sources with Gaussian noise, paving the way for more reliable and accurate subsurface imaging across diverse real conditions.
\end{abstract}

\paragraph{Keywords:} Full waveform inversion; Deep learning; Fourier-DeepONet; Benchmark datasets; Generalizability; Robustness

\section{Introduction}
\label{sec:intro}
Full waveform inversion (FWI) is a powerful tool used in the field of geophysics for imaging subsurface structures. It provides high-resolution images of geological formations, helping to enhance our comprehension of the Earth's crust and facilitating the identification of natural resources~\cite{virieux2009overview}. FWI is governed by partial differential equations (PDEs) and can be solved as a non-convex optimization problem. However, FWI remains a challenging optimization procedure due to its ill-posedness, nonlinearity, and high computational complexity~\cite{zhang2020fwi}.

With the development of deep learning techniques, data-driven approaches have been extensively studied in FWI for different neural network architectures, such as encoder-decoder-based convolutional neural networks (CNNs)~\cite{araya2018deep,wu2019inversionnet}, recurrent neural networks (RNNs)~\cite{sun2020theory,sun2021physics}, generative adversarial networks (GANs)~\cite{zhang2020data}, and so on. The utilization of data-driven FWI has been shown to provide accurate inversion results in an efficient manner~\cite{yu2021deep,lin2023physics}. An extensive overview of data-driven seismic inversion methods can be found in \citet{lin2023physics}.

The adoption of machine learning approaches for FMI presents various challenges. One such challenge pertains to the issue of generalization, which is the capacity of a machine learning model to exhibit competent performance on data that it has not previously encountered. Some prior research has proposed several approaches to address this challenge as listed below.
\begin{enumerate}
  \item \textbf{Build a large, diverse, and realistic dataset.} \citet{liu2021deep} generated a substantial quantity of realistic models containing dense layers, faults, and salt bodies through mathematical representation. In contrast, \citet{feng2021multiscale} extracted velocity maps from numerous natural images. \citet{alzahrani2022seismic} contributed to the development of velocity map training data by contrasting the generalization abilities of geologically inspired and purely geometric training sets. Meanwhile, \citet{dengopenfwi} investigated the correlation between dataset complexity and the generalization ability of the trained network.
  \item \textbf{Build a network with strong generalization ability.} \citet{zhang2020data} employed a generative model trained with 1-fault velocity maps and evaluated its efficacy in predicting 0-fault and 2-fault maps. \citet{aharchaou2020deep} constructed a U-net using data obtained from ocean-bottom nodes to extract low-frequency components, which could assist FWI in a wide-azimuth towed-steamer survey. \citet{li2022deep} developed a hybrid network that incorporates fully convolutional layers, an attention mechanism, and a residual unit to estimate velocity models from common source point gathers.
  \item \textbf{Incorporate governing physics into the purely data-driven method~\cite{karniadakis2021physics}.} \citet{ren2020physics} introduced a wave-equation-based forward modeling network cell for simulating seismic wave equations. \citet{jin2021unsupervised} and \citet{dhara2022elastic} established a connection between forward modeling and CNN in a loop, which transformed FWI into an unsupervised learning paradigm. \citet{zhu2022integrating} integrated neural networks into the optimization of FWI by representing the velocity model with a generative neural network.
\end{enumerate}
Although these methods prioritize the generalization of velocity models and networks, the generalization of source (and receiver components) has not yet been explored. The source wavelet is, however, an important component of the FWI process, a small disturbance in the source wavelet will lead to large discrepancies in the inverted model~\cite{sun2014source,lee2003source}. Hence, a good generalization of the source wavelet is crucial in a successful inversion. 

Another challenge of the data-driven FWI is model robustness, which is usually defined as the sensitivity of models to noise or outliers in the input data. Several studies have explored methods for improving the robustness of machine learning models. Miyato et al.~\cite{miyato2016adversarial} introduced adversarial training methods for semi-supervised text classification tasks, making use of the virtual adversarial training method, which is effective in improving the robustness of the model against adversarial perturbations and also leads to improved generalization performance. Lecuyer et al.~\cite{lecuyer2019certified} proposed a method for training models with provable guarantees on their robustness to adversarial examples, using techniques from the field of differential privacy. Hoffman et al.~\cite{hoffman2019robust} proposed Jacobian regularization, a technique that increases the classification margins of neural networks, thereby improving the robustness of the model. 
However, few studies of robustness focused on the field of data-driven FWI. Thus, more work is needed to understand the optimal strategies for improving the robustness of data-driven FWI models, and how to balance the trade-off between robustness and accuracy. Ultimately, the goal is to develop models that can reliably and accurately image subsurface structures across a range of different datasets and conditions.

In this work, we aim to address the aforementioned FWI challenges of generalization and robustness by developing a new machine learning approach. Our method is developed based on deep operator networks (DeepONets)~\cite{lu2021learning,jin2022mionet}, which have emerged as a powerful tool in computational science and engineering, as they offer a novel way to model complex physical systems and solve PDEs~\cite{lin2021operator,cai2021deepm,mao2021deepm,lu2022multifidelity,deng2022approximation,jiang2023fourier,mao2023ppdonet}. A DeepONet comprises two constituent sub-networks: a ``trunk'' network and a ``branch'' network. Recent works~\cite{lu2022comprehensive, di2023neural, zhu2023reliable} showed that the DeepONet has satisfactory robustness against inputs with noise.
Taking linear instability waves in high-speed boundary layers as an example, when Gaussian noise of 0.1\% standard deviation is added to the input during testing, the error of DeepONet is almost unchanged, while the error of CNN experiences a substantial increase, by two orders of magnitude~\cite{lu2022comprehensive}. Even if Gaussian noise is up to 1\%, the error of DeepONet only experiences a minor increase ($<2\%$)~\cite{di2023neural}. This example demonstrates the better robustness of DeepONet over traditional neural networks.

The vanilla DeepONet uses an inner product as the decoding mechanism to construct the output~\cite{lu2021learning}, which results in blurred outputs of subsurface structures as observed in our experiments. Here, we incorporate more sophisticated networks as decoders, facilitating the production of clearer and more accurate predictions. Specifically, we develop a Fourier-enhanced DeepONet (Fourier-DeepONet) by utilizing U-FNO~\cite{wen2022u} (a block combining Fourier neural operator~\cite{li2020fourier} and U-Net~\cite{ronneberger2015u}) as the decoder. In addition, in contrast to the vanilla DeepONet, whose trunk net utilizes output function coordinates as input, we propose employing source parameters as the inputs of the trunk net. By adopting this approach, we facilitate the data-driven model of full waveform inversion with sources of variable frequencies and locations. To validate its performance, we develop three new FWI benchmark datasets (FWI-F, FWI-L, and FWI-FL) with varying sources. In comparison to existing data-driven FWI techniques, such as InversionNet~\cite{wu2019inversionnet} and VelocityGAN~\cite{zhang2020data}, Fourier-DeepONet exhibits a considerably enhanced accuracy and generalizability. Moreover, our experiments show that Fourier-DeepONet is much more robust when confronted with inputs containing noise or absent traces.

The paper is organized as follows. In Section~\ref{sec:fwi&datasets}, we first introduce the problem setup of FWI and the OpenFWI benchmark dataset, and then we generate three new benchmark datasets (FWI-F, FWI-L, and FWI-FL) with varying source frequencies and locations. In Section~\ref{sec:methods}, after briefly introducing the vanilla DeepONet, we propose Fourier-DeepONet in parameter spaces. In Section~\ref{sec:results}, we compare the performance of Fourier-DeepONet and two baseline models for the three collections of benchmark datasets. Section~\ref{sec:conclusions} summarizes the paper and discusses some future directions.

\section{Full waveform inversion}
\label{sec:fwi&datasets}

We first introduce the problem setup of seismic FWI and forward modeling in Section~\ref{subsec:fwi} and then present the datasets in Section~\ref{sec:dataset}.

\subsection{Problem setup}
\label{subsec:fwi}

The acoustic wave equation is a second-order PDE that describes the propagation of sound waves or pressure waves in a fluid or solid medium. The following is the governing equation of acoustic waves in an isotropic medium with a uniform density:
\begin{equation*}
    \nabla^2 p(\textbf{r}, t) - \frac{1}{c(\textbf{r})^2}\frac{\partial^2p(\textbf{r}, t)}{\partial t^2} = s(\textbf{r}, t),    
\end{equation*}
where $t$ is time, $\textbf{r}$ the spatial location, $s$ is the source function, $p$ is the pressure wavefield, and $c$ is the velocity map of the subsurface medium. In this work, we focus on two-dimensional acoustic wave equation, i.e.,
\begin{equation*}
    \nabla^2 p(x, z, t) - \frac{1}{c(x, z)^2}\frac{\partial^2p(x, z, t)}{\partial t^2} = s(x, z, t),    
\end{equation*}
where $x$ is horizontal location (length), and $z$ is vertical location (depth). Here, we use a point source near the surface at $z=10$ m generated by Ricker wavelet, and the expression of source function is
\begin{equation}
\label{equ:source}
s(x, z, t)=s_0(t; f)\delta(x-x_s, z-10),
\end{equation}
where
$$s_0(t; f) = (1-2\pi^2 f^2 t^2)e^{-\pi^2 f^2 t^2}$$
is the amplitude of the Ricker wavelet (Fig.~\ref{fig:source_noise}B) with frequency $f$, $\delta$ represents the Dirac delta function, and $x_s$ is the predetermined horizontal location of the source.

We utilize the seismic forward modeling algorithm~\cite{moczo2007finite} to solve pressure wavefield from velocity maps, which utilizes finite difference methods with zero initial conditions. Additionally, the absorbing boundary~\cite{engquist1977absorbing} condition is employed to account for the wave attenuation and prevent wave reflection at the edges of the simulation domain. We use $g$ to denote the forward modeling operator, i.e.,
\begin{equation*}
    p = g(c; s).
\end{equation*}

\begin{figure}[htbp]
    \centering
    \includegraphics[width=\textwidth]{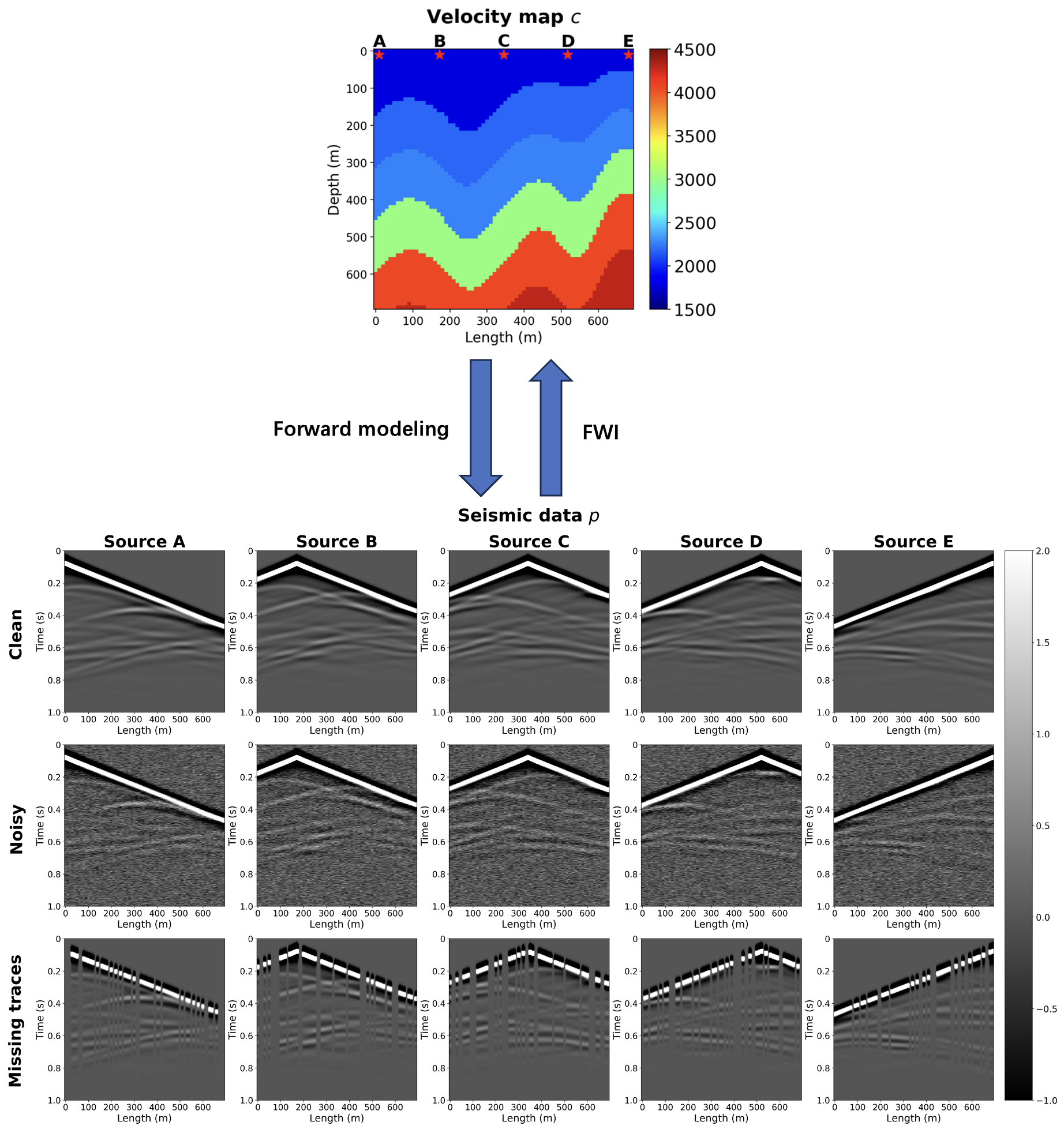}
    \caption{\textbf{Schematic illustration of forward modeling and FWI.} Forward modeling generates seismic data from velocity maps, while the purpose of FWI is to infer velocity maps from seismic data measurements. The five red stars in the velocity map are the five point sources utilized to generate seismic data. The seismic data can be clean, noisy, or with missing traces.}
    \label{fig:fwi}
\end{figure}

In contrast, FWI aims to derive velocity maps $c$ from pressure wavefield $p$. In practice, however, only a limited number of receivers can be used to acquire pressure wavefield, and the receivers are typically placed on the surface (i.e., $z=0$). To compensate for the missing subsurface pressure wavefield, we usually consider having the surface pressure wavefield data for multiple sources. Specifically, we assume that we have five source functions $\{s_\text{A}, s_\text{B}, s_\text{C}, s_\text{D}, s_\text{E}\}$ at different locations (Fig.~\ref{fig:fwi}), and for a source function $s_i$ with $i \in \{A, B, C, D, E\}$, the corresponding pressure wavefield is $p_i = g(c; s_i)$. We denote the pressure wavefield at surface by $p_{i,0}(x, t)=p_i(x, z=0, t)$. Since $p_{i,0}(x, t)$ is only the data collected by receivers on the surface, we call $p_{i,0}(x, t)$ seismic data instead of pressure wavefield in the rest of the paper. Then, the problem of FWI is defined as learning the mapping
\begin{equation*}
    \left[ p_{\text{A},0},\ p_{\text{B},0},\ p_{\text{C},0},\ p_{\text{D},0},\ p_{\text{E},0} \right] \mapsto c.
\end{equation*}

Fig.~\ref{fig:fwi} is a schematic illustration of forward modeling and FWI. The five red stars on the top of the velocity map are the five point sources utilized to generate seismic data. Each source is corresponding to one column of seismic data in Fig.~\ref{fig:fwi}. The resolution of seismic data in space is 10 m, i.e., the distance between two adjacent receivers is 10 m. The resolution in time is 0.001 second, and the total time is one second. In this study, we not only consider clean seismic data, but also noisy seismic data and seismic data with missing traces, as shown in Fig.~\ref{fig:fwi}.

In the rest of the paper, the figures of seismic data and velocity maps will carry the same labels and colorbars as presented in Fig~\ref{fig:fwi}. For the sake of conciseness, these elements will not be included in subsequent figures.

\subsection{Datasets}
\label{sec:dataset}

The availability of large-scale datasets with a variety of subsurface structures provides a comprehensive testbed for evaluating and enhancing machine learning models for FWI. We first introduce an existing collection of benchmark datasets OpenFWI in Section~\ref{subsec:openfwi} and then generate three new collections of benchmark datasets (FWI-F, FWI-L, and FWI-FL) in Section~\ref{subsec:fwi-fl}.

\subsubsection{OpenFWI}
\label{subsec:openfwi}

\begin{figure}[htbp]
    \centering
    \includegraphics[width=.9\textwidth]{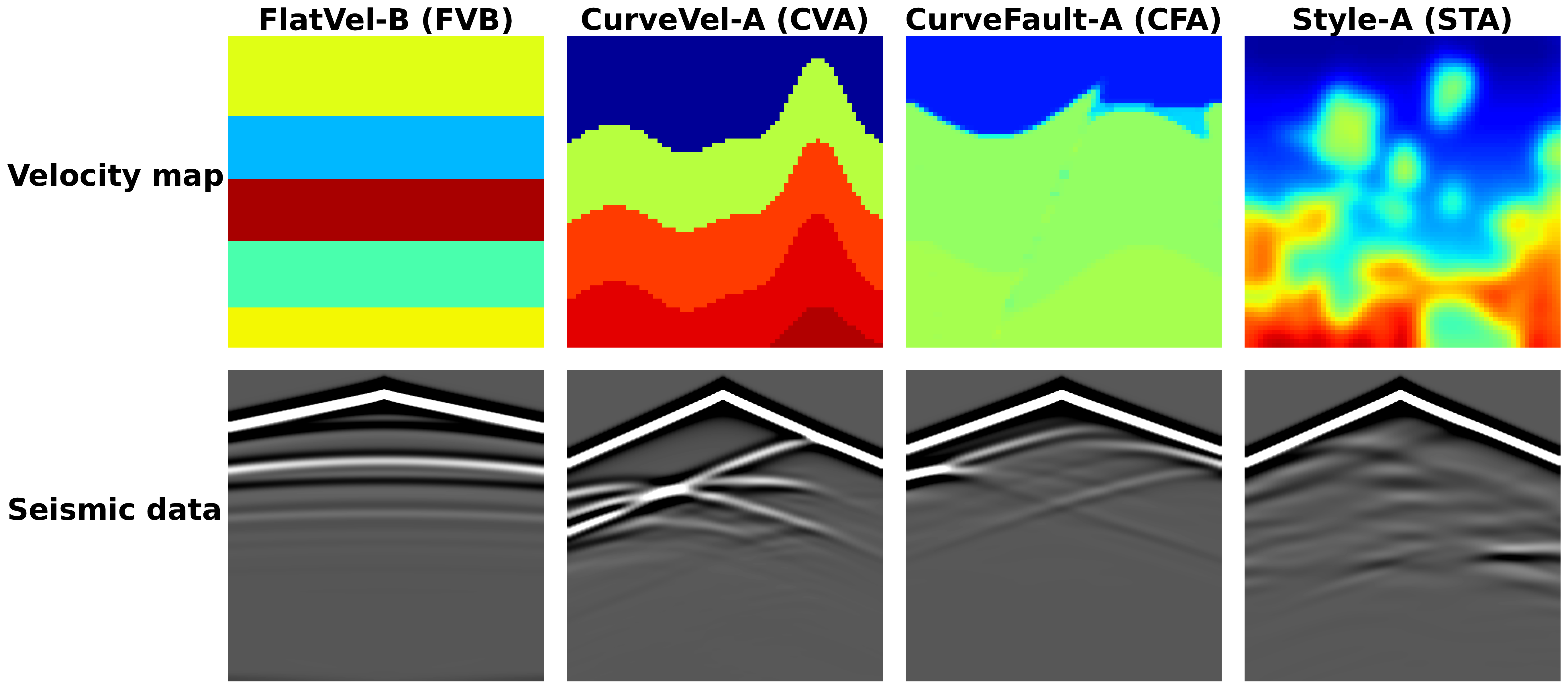}
    \caption{\textbf{Examples of velocity maps (top row) and seismic data (bottom row) for FVB, CVA, CFA and STA datasets.} The seismic data comes from the source C in Fig.~\ref{fig:fwi}.}
    \label{fig:datasets}
\end{figure}

OpenFWI~\cite{dengopenfwi} is a collection of benchmark datasets on data-driven seismic FWI. In this study, we choose velocity maps of four datasets from OpenFWI: ``FlatVel-B'' (FVB), ``CurveVel-A'' (CVA), ``CurveFault-A'' (CFA), and ``Style-A'' (STA), see examples in Fig.~\ref{fig:datasets}.
\begin{itemize}
    \item The FVB dataset is designed for situations with flat layers and clear interfaces, representing relatively simple scenarios, where layers are horizontally oriented without much variation. 
    \item The CVA dataset is intended for situations involving curved layers with distinct interfaces. This dataset poses greater complexity compared to the FVB dataset, as the presence of curved layers creates further complications in the process of subsurface imaging. 
    \item The CFA dataset is developed for scenarios that involve curved layers and discontinuities resulting from faults in velocity maps. The existence of faults increases the intricacy of the subsurface imaging process, making it more difficult to acquire precise velocity maps.
    \item The STA dataset is generated through style transfer techniques applied to natural images, signifying that the dataset draws inspiration from actual geological formations. This dataset aims to offer a more authentic depiction of subsurface imaging situations.
\end{itemize}
The number of training/testing cases for FVB, CVA, CFA, and STA are 24K/6K, 24K/6K, 48K/6K, and 60K/7K, respectively.



\subsubsection{FWI-F, FWI-L, and FWI-FL}
\label{subsec:fwi-fl}

According to Eq.~\eqref{equ:source}, the Ricker wavelet source has two main parameters: frequency $f$ and horizontal location $x_s$. OpenFWI adopts fixed values for all parameters: the source frequencies are fixed at 15 Hz, and the five sources are uniformly distributed on the surface, whose horizontal coordinates are 0, 172.5, 345, 517.5, and 690 m. Here, we generate three new datasets that vary in terms of source frequencies, locations, or both.
\begin{itemize}
    \item FWI-F: The source frequency is ranging from 5 to 25 Hz.
    \item FWI-L: We allow the middle three sources (B, C, and D) to move up to 50 m either to the left or the right, while the sources at the boundaries (A and E) can move towards the center. Hence, the possible horizontal coordinates of the five sources are [0, 50], [122.5, 222.5], [295, 295], [467.5, 567.5], and [640, 690]~m, and the source spacing varies between 72.5 and 272.5 m.
    \item FWI-FL: The source frequency and location follow the patterns of FWI-F and FWI-L, respectively.
\end{itemize}
The number of training/testing cases is the same as OpenFWI. Table~\ref{tab:config} displays the source frequencies and locations for the four datasets.

\begin{table}[htbp]
\centering
\caption{\textbf{Source frequencies and locations of different datasets.} The source frequency and location in OpenFWI are fixed, whereas in FWI-F, FWI-L, and FWI-FL, the source frequencies, locations, and both frequencies and locations vary, respectively.}
\label{tab:config}
\begin{tabular}{c|c|ccccc}
\toprule
\multirow{2}{*}{Dataset} & Frequency & \multicolumn{5}{c}{Location (m)}\\
 & (Hz) & Source A & Source B & Source C & Source D & Source E\\
\midrule
OpenFWI & 15 & 0 & 172.5 & 345 & 517.5 & 690 \\
FWI-F & [5, 25] & 0 & 172.5 & 345 & 517.5 & 690 \\
FWI-L & 15 & [0, 50] & [122.5, 222.5] & [295, 295] & [467.5, 567.5] & [640, 690]\\
FWI-FL & [5, 25] & [0, 50] & [122.5, 222.5] & [295, 295] & [467.5, 567.5] & [640, 690]\\
\bottomrule
\end{tabular}
\end{table}

\section{Methods}
\label{sec:methods}


Traditional deep learning methods generally focus on finite-dimensional data, such as images or time series, while DeepONet aims to learn mappings between infinite-dimensional function spaces~\cite{lu2021learning}. From the standpoint of computation, this operator regression approach provides the merit of simulating complex nonlinear systems without requiring neural network retraining. Once the DeepONet has been trained, it can be employed on new input functions, thereby yielding results considerably more rapidly than conventional numerical solvers. Moreover, the DeepONet exhibits adaptability, as it can be applied to simulation data, experimental data, or a fusion of both types. This flexibility allows for the incorporation of experimental data covering a broad range of spatio-temporal scales, thus empowering researchers to estimate dynamics more precisely by synthesizing existing datasets~\cite{higgins2021generalizing}.

\subsection{Vanilla DeepONet}

We denote the input function by $v$ defined on the domain $D \subset \mathbb{R}^d$
$$v: x \mapsto v(x), \quad x \in D,$$
and denote the output function by $u$ defined on the domain \(D' \subset \mathbb{R}^{d'}\)
$$u: y \mapsto u(y), \quad y \in D'.$$
Let $\mathcal{V}$ and $\mathcal{U}$ be the spaces of $v$ and $u$, respectively. Then, the mapping
from the input function $v$ to the output function $u$ is denoted by an operator
$$\mathcal{G}: \mathcal{V}\ni v \mapsto u \in \mathcal{U}.$$

The DeepONet~\cite{lu2021learning} is a neural network architecture, tailored specifically to approximate nonlinear operators between functional spaces. A vanilla DeepONet consists of two separate networks: the ``trunk'' network that accepts the coordinates $y \in D'$ as input; and the ``branch'' network that uses a discretized function $v$ as its input. We particularly assess $v$ at $m$ predetermined positions $\left\{x_1, x_2, x_3, \dots, x_m\right\}$ to acquire the pointwise evaluations $\textbf{v}= \left\{v(x_1), v(x_2), v(x_3), \dots, v(x_m)\right\}$.
Then the network's output is presented as
$$u(y) \approx \mathcal{G}(\textbf{v})(y)=\sum_{i=1}^{r}b_i(\textbf{v})t_i(y)+b_0$$
where $\big\{b_1, b_2,...,b_r \big\}$ and $\big\{t_1, t_2,...,t_r \big\}$ are the $r$ outputs of the branch net and trunk net, respectively, and $b_0 \in \mathbb{R}$ is a bias.

\subsection{Fourier-DeepONet in parameter spaces}
\label{subsec:fourier_deepONet}
In the context of FWI applications, the branch network is tailored to receive seismic data as input. Unlike the trunk network of vanilla DeepONet, which employs coordinates of output functions as input, we propose utilizing source parameters as input for the trunk network. Consequently, our DeepONet is established in the parameter space rather than the spatial-temporal domain. We denote the seismic data by $\textbf{p}=\left[ p_{\text{A},0},\ p_{\text{B},0},\ p_{\text{C},0},\ p_{\text{D},0},\ p_{\text{E},0} \right]$, and source parameters by $\boldsymbol{\xi}$ ($\boldsymbol{\xi}=f$ or $\boldsymbol{\xi}$=[$x_{s_{\text{A}}}$, $x_{s_{\text{B}}}$, $x_{s_{\text{C}}}$, $x_{s_{\text{D}}}$, $x_{s_{\text{E}}}$]), and then the problem of FWI with varying source parameters is defined as learning the mapping
$$\mathcal{G}: (\textbf{p}, \boldsymbol{\xi}) \mapsto c.$$

First, we need to discretize the input function within a finite-dimensional space. We particularly assess each $p_{i,0}$ at $R$ predetermined receiver positions at the surface $\left\{x_1, x_2, x_3, \dots, x_R\right\}$ to acquire the pointwise evaluations, and each receiver records at $T$ times $\left\{t_1, t_2, t_3, \dots, t_T\right\}$ in the wavefields. Here, $R=70$, and $T=1000$. Thus, discretized seismic data for each source $i \in \{A, B, C, D, E\}$ is a matrix of shape (1000, 70), and the discretized seismic data $\textbf{p}$ is a tensor of shape (1000, 70, 5). The source parameter $\boldsymbol{\xi}$ is a vector that has a length of 1 for source frequencies, a length of 5 for source locations, or a length of 6 for both source frequencies and locations. The Fourier-DeepONet output (i.e., velocity maps) is a function on the spatial domain. Thus, $c$ needs also to be discretized within a finite-dimensional space. We particularly assess $c$ at $W \times H$ predetermined positions, where $W=70$ and $H=70$ correspond to the horizontal and vertical extents of velocity maps. Thus,
the discretized $\textbf{c}$ is a matrix of shape (70, 70).

\begin{figure}[htbp]
    \centering
    \includegraphics[width=\textwidth]{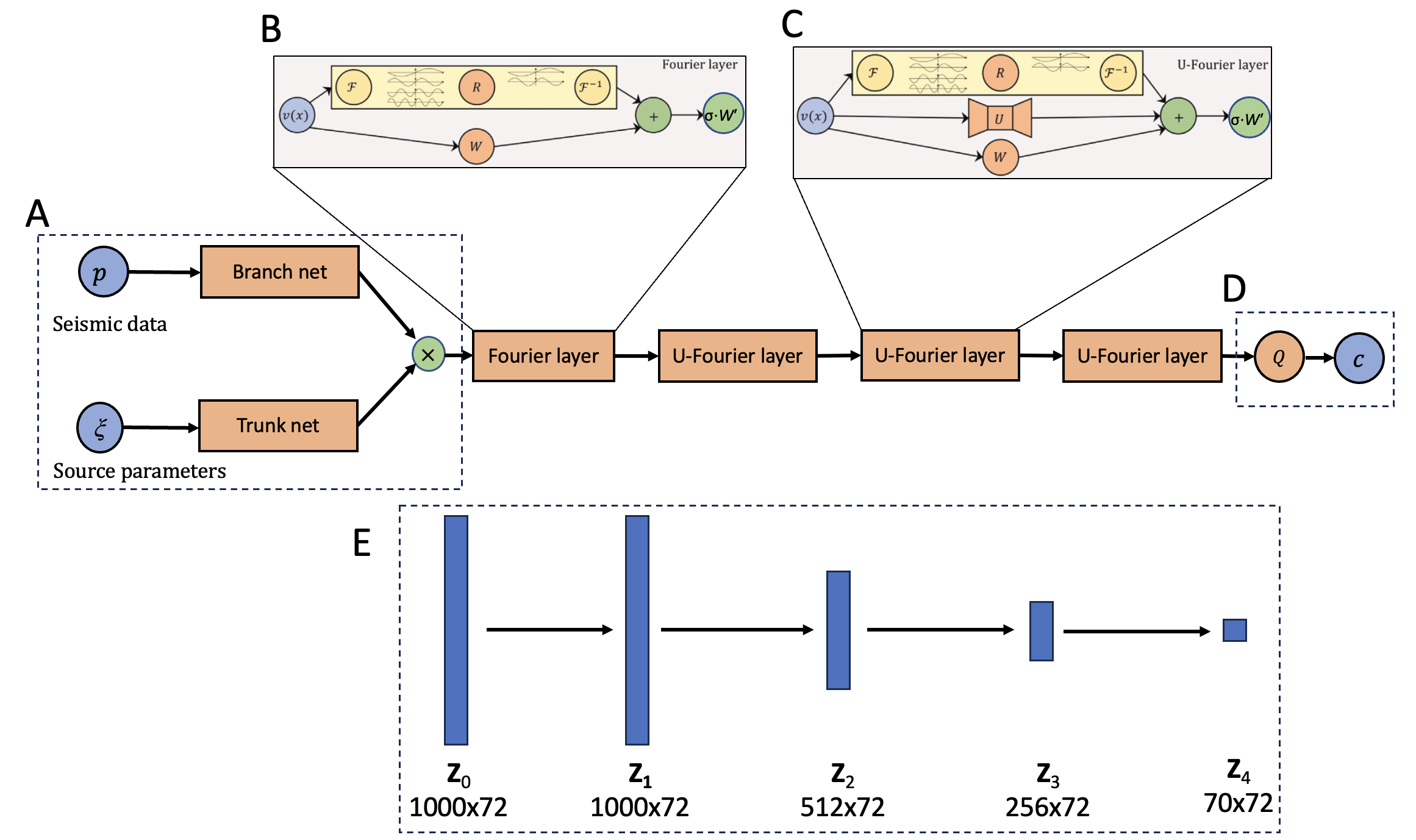}
    \caption{\textbf{Fourier-DeepONet architecture.} (\textbf{A}) Branch net and trunk net are two linear transformations lifting inputs to high dimensional space. Green circle represents the merger operation which denotes point-wise multiplication. (\textbf{B}) Fourier layer, adapted from~\cite{li2020fourier}. (\textbf{C}) U-Fourier layer, adapted from~\cite{wen2022u}. (\textbf{D}) Projection layer $Q$. (\textbf{E}) Shapes of outputs $\textbf{z}_i$ in one channel, $i \in \{0, 1, 2, 3, 4\}$.}
    \label{fig:architecture}
\end{figure}

In Fourier-DeepONet, the branch net and trunk net encode the seismic data $\textbf{p}$ and parameters $\boldsymbol{\xi}$, respectively (Fig.~\ref{fig:architecture}A). We denote outputs of the branch and trunk nets by $\textbf{b}$ and $\textbf{t}$: 
$$\textbf{b} = B(\textbf{p}) \in \mathbb{R}^{T \times R \times C},$$
$$\textbf{t} = T(\boldsymbol{\xi}) \in \mathbb{R}^{C},$$
where $C$ is the number of channels, and the branch net $B$ and trunk net $T$ are two linear transformations to increase the number of channels to $C$. Here, $C=64$. A merger operation is needed to merge branch and trunk outputs together. We compare three possible merger operations in Section~\ref{subsec:operation}. Here, we choose pointwise multiplication after tensor broadcasting as the merger operation:
$$
     \textbf{z}_0 = \textbf{b} \odot \textbf{t},
$$
where $\textbf{z}_0$ is the output of merger operation with the dimension of $T \times R \times C$. We need to transform the dimension from $T \times R \times C$ to $R \times R \times C$ to generate velocity maps $c$. However, for both the Fourier layer and U-Net layer, the dimension of input and output are the same. To solve this issue, we add a linear layer before the activation layer in each Fourier and U-Fourier layer.

Next, we utilize one Fourier layer (Fig.~\ref{fig:architecture}B) and three U-Fourier layers (Fig.~\ref{fig:architecture}C) as the merger net:
$$
\textbf{z}_1 =\sigma \left(W'_1 \left(\mathcal{F}^{-1}\left(R_1 \cdot \mathcal{F}\left(\textbf{z}_0\right)\right)+W_1 \cdot \textbf{z}_0+b_1\right)\right),
$$
$$
\textbf{z}_2 =\sigma \left(W'_2 \left(\mathcal{F}^{-1}\left(R_2 \cdot \mathcal{F}\left(\textbf{z}_1\right)\right)+\mathcal{U}_2\left(\textbf{z}_1\right)+W_2 \cdot \textbf{z}_1+b_2\right)\right),
$$
$$
\textbf{z}_3 =\sigma \left(W'_3 \left(\mathcal{F}^{-1}\left(R_3 \cdot \mathcal{F}\left(\textbf{z}_2\right)\right)+\mathcal{U}_3\left(\textbf{z}_2\right)+W_3 \cdot \textbf{z}_2+b_3\right)\right),
$$
$$
\textbf{z}_4 =\sigma \left(W'_4 \left(\mathcal{F}^{-1}\left(R_4 \cdot \mathcal{F}\left(\textbf{z}_3\right)\right)+\mathcal{U}_4\left(\textbf{z}_3\right)+W_4 \cdot \textbf{z}_3+b_4\right)\right),
$$
where $\textbf{z}_1$, $\textbf{z}_2$, $\textbf{z}_3$, and $\textbf{z}_4$ are corresponding to the outputs of Fourier, U-Fourier 1, U-Fourier 2, and U-Fourier 3 in Table~\ref{table: architecture}. Here, $\mathcal{F}$ is two-dimensional Fast Fourier Transform (FFT), $\mathcal{F}^{-1}$ is inverse two-dimensional FFT, $\mathcal{U}$ is an U-Net layer, $R$, $W$ and $W'$ are weight matrices, and $b$ is a bias. For more details about Fourier layers and U-Fourier layers, see Refs.~\cite{wen2022u, li2020fourier, lu2022comprehensive, ronneberger2015u}. We discuss other configurations of Fourier and U-Fourier layers in Section~\ref{subsec:fno_ufno} and find  the combination of one Fourier layer and three U-Fourier layers is a better choice given the balance of performance and efficiency.
In the end, a non-linear transformation $Q$ (Fig.~\ref{fig:architecture}D) is applied to project output of the last U-Fourier layer $\textbf{z}_4$ to velocity map $c$.

We note that to ensure that the shape is compatible with U-Net, at the beginning of the branch net, we use an extra padding layer to convert the dimension from $1000 \times 70 \times 64$ to $1000 \times 72 \times 64$. More details of the operations and shapes of Fourier-DeepONet are summarized in Table~\ref{table: architecture}.

\begin{table}[htbp]
    \centering
    \caption{\textbf{Fourier-DeepONet architecture.} The merger net includes one Fourier layer and three U-Fourier layers. The output shape does not include the batch size dimension.}
    \label{table: architecture}
    \begin{tabular}{lll}
    \toprule 
     & Operations & Output shape\\
    \midrule
    Branch net & Padding, Linear & (1000, 72, 64)\\  
    Trunk net & Linear & (64)\\
    Merger operation & Pointwise multiplication of branch and trunk outputs & (1000, 72, 64)\\
    \midrule 
    Fourier & Add(Fourier2d+Conv1d), ReLU & (1000, 72, 64)\\
    U-Fourier 1 & Add(Fourier2d+Conv1d+UNet2d), Linear, ReLU & (512, 72, 64)\\
    U-Fourier 2 & Add(Fourier2d+Conv1d+UNet2d), Linear, ReLU & (256, 72, 64)\\
    U-Fourier 3 & Add(Fourier2d+Conv1d+UNet2d), Linear, ReLU & (70, 72, 64)\\
    Projection & Linear, ReLU, Linear, Slicing & (70, 70)\\
    \bottomrule
    \end{tabular}
\end{table}

\subsection{Baseline models}
\label{subsec:baseline}

To compare with Fourier-DeepONet, we present two widely-used data-driven FWI models, InversionNet~\cite{wu2019inversionnet} and VelocityGAN~\cite{zhang2020data}, as baseline models.

InversionNet employs a fully-convolutional network composed of an encoder and a decoder to model the seismic inversion process. The encoder accepts 2D seismic data as input, represented as a matrix of time samples and receiver locations. The decoder generates a 2D velocity map, illustrating subsurface rock properties at varying depths and locations. The network is trained using supervised learning on a dataset containing paired seismic data and corresponding velocity maps. InversionNet has proven to be an efficient method for seismic inversion, achieving state-of-the-art results on multiple benchmark datasets~\cite{dengopenfwi, zhang2020data, jin2021unsupervised}.

VelocityGAN is based on the idea of generative adversarial networks (GANs) and comprises a generator and a discriminator. The generator features an encoder-decoder architecture, similar to InversionNet, which serves as a surrogate model for FWI. The discriminator, a CNN, is trained to differentiate between genuine and counterfeit velocity maps. The generator is trained to create velocity maps capable of deceiving the discriminator, promoting more accurate and realistic outcomes. Moreover, VelocityGAN also utilizes network-based deep transfer learning to enhance the model's robustness and generalization. This method involves fine-tuning a pre-trained model for a related task and employing it as the initial point for GAN training. This approach allows the model to learn more rapidly and efficiently by leveraging the knowledge already embedded in the pre-trained model.

\section{Results}
\label{sec:results}

In this section, we test the proposed Fourier-DeepONet on FWI with varying source frequencies (Section~\ref{subsec:fre}), locations (Section~\ref{subsec:dev}), or both (Section~\ref{subsec:f_loc}). We then evaluate the model robustness in Section~\ref{subsec:robustness} and discuss the architecture of Fourier-DeepONet in Section~\ref{subsec:architecture}.

The seismic data and velocity maps are normalized to range between -1 and 1 during training and testing. To evaluate the performances of different models, we adopt the following metrics~\cite{dengopenfwi}: (1) mean absolute error (MAE), (2) root mean squared error (RMSE), (3) structural similarity (SSIM), and (4) $L^2$ relative error. MAE and RMSE are two commonly used evaluation metrics in regression problems, which quantify the differences between predicted values and actual values. MAE is calculated as the mean of the absolute differences between the predicted and actual values. RMSE is calculated as the square root of the mean of the squared differences between the predicted and actual values. By squaring the errors, it emphasizes larger errors compared to MAE. SSIM measures the similarity between two images based on how closely they match from a perceptual perspective~\cite{wang2004image}, providing a representation of how humans perceive image quality. The $L^2$ relative error is a metric used to measure the difference between predictions and ground truth by taking into account the magnitudes of the values. Smaller MAE, RMSE, and $L^2$ relative error, or larger SSIM indicate better model performance.

\subsection{Source frequencies}
\label{subsec:fre}

The source frequency plays a crucial role in FWI, as it directly impacts the resolution and convergence of the subsurface velocity model. The source frequency determines the frequency content of the seismic wavelet used in the modeling process, which in turn affects the accuracy and stability of the inversion. Higher source frequencies lead to shorter wavelengths, which in turn provide higher-resolution images of the subsurface. This means that finer details and smaller geological features can be resolved when using higher frequencies. However, higher frequencies are also more susceptible to attenuation and noise, making it challenging to acquire high-quality data at higher frequencies. Moreover, low-frequency components of the source wavelet are essential for the overall convergence of the FWI process. Low frequencies help in building a macro velocity model that captures the large-scale structure of the subsurface, which acts as a good starting point for the inversion. If the source lacks low-frequency content, the inversion process may get trapped in local minima, leading to inaccurate results. Therefore, the choice of source frequency in FWI is critical for achieving accurate subsurface velocity models.

The frequency content should be carefully considered to balance the trade-off between resolution and stability while ensuring the convergence of the inversion process. Despite the significance of source frequency in FWI, OpenFWI and most of other FWI datasets use a fixed source frequency. This simplification may not accurately represent real-world FWI problems, as it overlooks the complexities and challenges associated with varying source frequencies, which limit the generalizability of the models trained on such datasets. To better prepare data-driven FWI models for real-world scenarios, it is important to develop datasets and training strategies that consider varying source frequencies. This will ensure that the trained models can handle a broader range of FWI problems, improving their robustness and applicability in practical situations.

\subsubsection{Comparison with pretrained baseline models}

We use the dataset FWI-F with varying source frequencies introduced in Section~\ref{subsec:fwi-fl}. The parameters of FWI-F remain the same as in OpenFWI, except that the source frequencies are randomly distributed between 5 Hz and 25 Hz. We train Fourier-DeepONet on FWI-F, while we use the baseline models (InversionNet and VelocityGAN) pretrained on OpenFWI from Ref.~\cite{dengopenfwi}.

\begin{figure}[htbp]
    \centering
    \includegraphics[width=\textwidth]{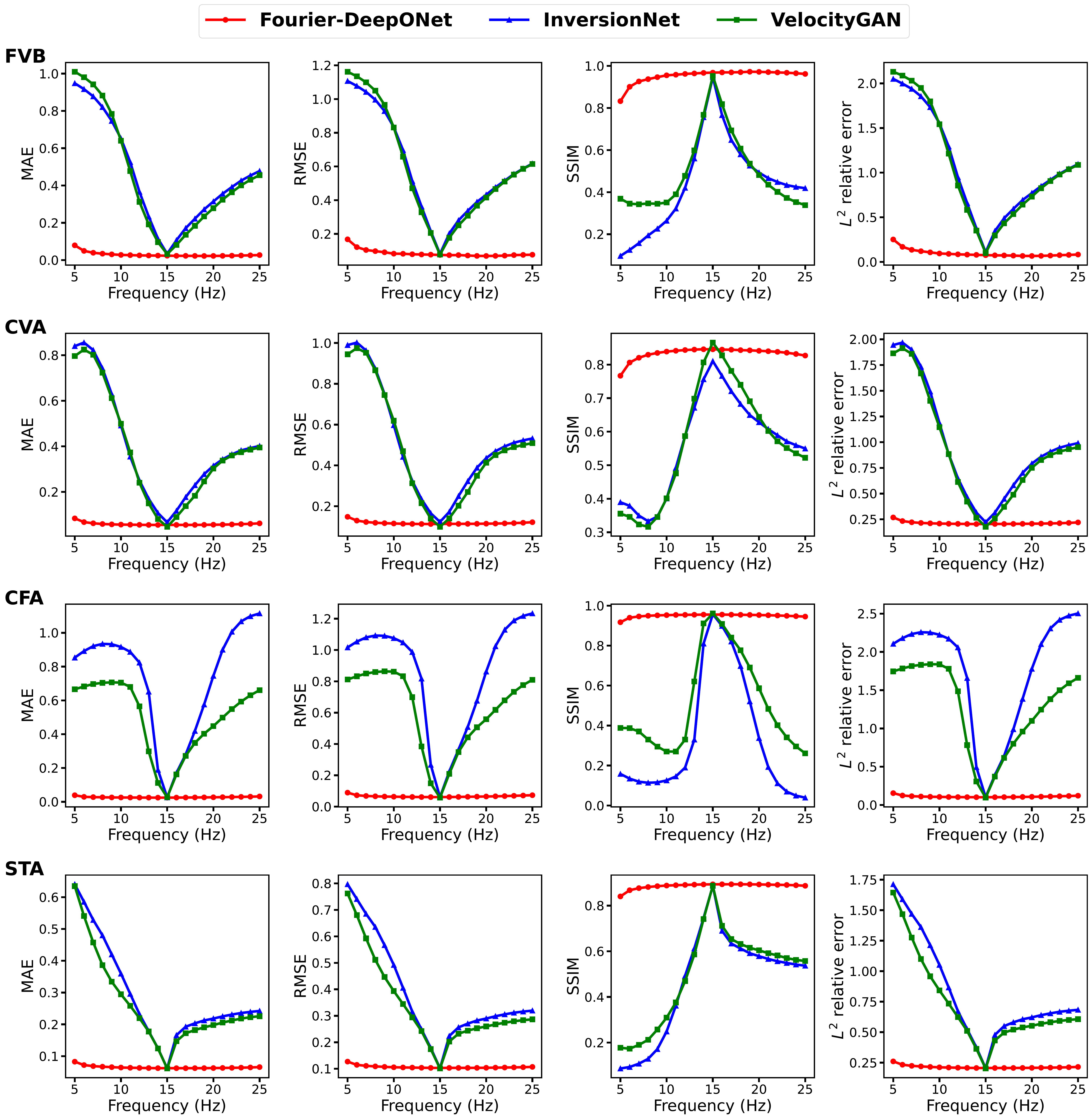}
    \caption{\textbf{Performance of three methods on four datasets (FVB, CVA, CFA, and STA) of different sources frequencies.} Fourier-DeepONet is trained on FWI-F; InversionNet and VelocityGAN are trained on OpenFWI. Fourier-DeepONet performs better for all datasets across a wide range of frequencies, while InversionNet and VelocityGAN can only give accurate predictions when the frequency is 15 Hz.}
    \label{fig:frequency}
\end{figure}

We compare the performance (MAE, RMSE, SSIM, and $L^2$ relative error) of Fourier-DeepONet, InversionNet, and VelocityGAN in FVB, CVA, CFA, and STA datasets in Fig.~\ref{fig:frequency}. Fourier-DeepONet performs well for all datasets across a range of frequencies (Fig.~\ref{fig:frequency}, red lines). However, InversionNet and VelocityGAN (Fig.~\ref{fig:frequency}, blue and green lines) can only give accurate predictions when the frequency is 15 Hz. When the frequency deviates from 15 Hz, their performance is much poorer than Fourier-DeepONet. For example, for the FVB dataset with the source frequency as 15 Hz, the $L^2$ relative errors of the three models are around 10\%. However, when we reduce the source frequency to 10 Hz, the $L^2$ relative errors of InversionNet and VelocityGAN increase significantly to 150\%, while the $L^2$ relative error of Fourier-DeepONet is still about 10\%. Thus, Fourier-DeepONet can accurately predict velocity maps for frequencies ranging from 5 Hz to 25 Hz, while InversionNet and VelocityGAN can only predict velocity maps for a fixed frequency, indicating their weak generalizability as also discussed in Refs.~\cite{wu2019inversionnet,zhang2020data}.

We find that the case of higher frequency is easier to learn than the case of lower frequency (Fig.~\ref{fig:frequency}). Specifically, for InversionNet and VelocityGAN, the error degrades faster when the frequency decreases than when the frequency increases. Similarly, for Fourier-DeepONet, the error is larger when the frequency is at 5 Hz than when the frequency is at 25 Hz. In addition, for InversionNet and VelocityGAN, when the frequency deviates significantly from 15 Hz (e.g., smaller than 10 Hz or larger than 25Hz for CFA), the error no longer increases, since the error is already very large ($L^2$ relative error $\sim$200\%).

\begin{figure}[htbp]
    \centering
    \includegraphics[width=\textwidth]{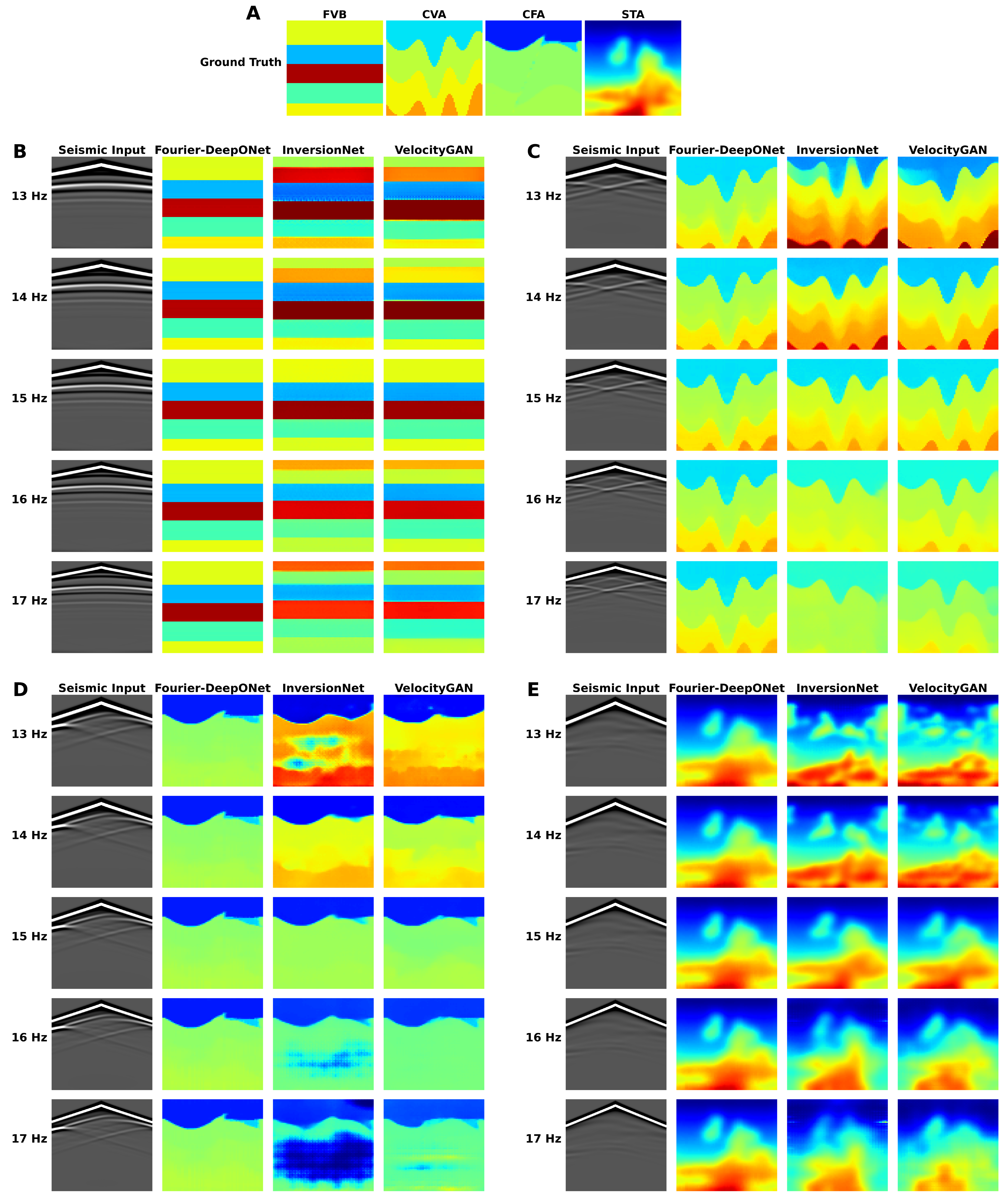}
    \caption{\textbf{Examples of velocity maps predicted by three methods on four datasets with source frequencies from 13 to 17 Hz.} (\textbf{A}) Ground truth of examples from FVB, CVA, CFA, and STA datasets. (\textbf{B}) Predictions for the FVB case. (\textbf{C}) Predictions for the CVA case. (\textbf{D}) Predictions for the CFA case. (\textbf{E}) Predictions for the STA case. The seismic data comes from the source C in Fig.~\ref{fig:fwi}.}
    \label{fig:frequency_heatmap}
\end{figure}

We show several examples of velocity maps predicted by the three models on FVB, CVA, CFA and STA datasets in Fig.~\ref{fig:frequency_heatmap}. When the frequency is at 15 Hz, all three methods yield satisfactory predictions. However, when the frequency deviates slightly from 15 Hz, InversionNet and VelocityGAN show inaccurate predictions. When frequency increases, both InversionNet and VelocityGAN tend to underestimate values of velocity maps. In contrast, when frequency decreases, both InversionNet and VelocityGAN tend to overestimate values of velocity maps. Fourier-DeepONet always has accurate predictions in all cases.

Another serious issue of InversionNet and VelocityGAN is that their predictions may have an incorrect number of layers. For example, in Fig.~\ref{fig:frequency_heatmap}B, predictions of InversionNet and VelocityGAN split into two separate layers at the top layer if the frequency is not at 15 Hz.

\subsubsection{Comparison with improved baseline models}

For the results described above, we utilize the pretrained InversionNet and VelocityGAN based on the OpenFWI datasets, while Fourier-DeepONet is trained with FWI-F. Here, we improve the two baseline models by training them with FWI-F.

\begin{figure}[htbp]
    \centering
    \includegraphics[width=\textwidth]{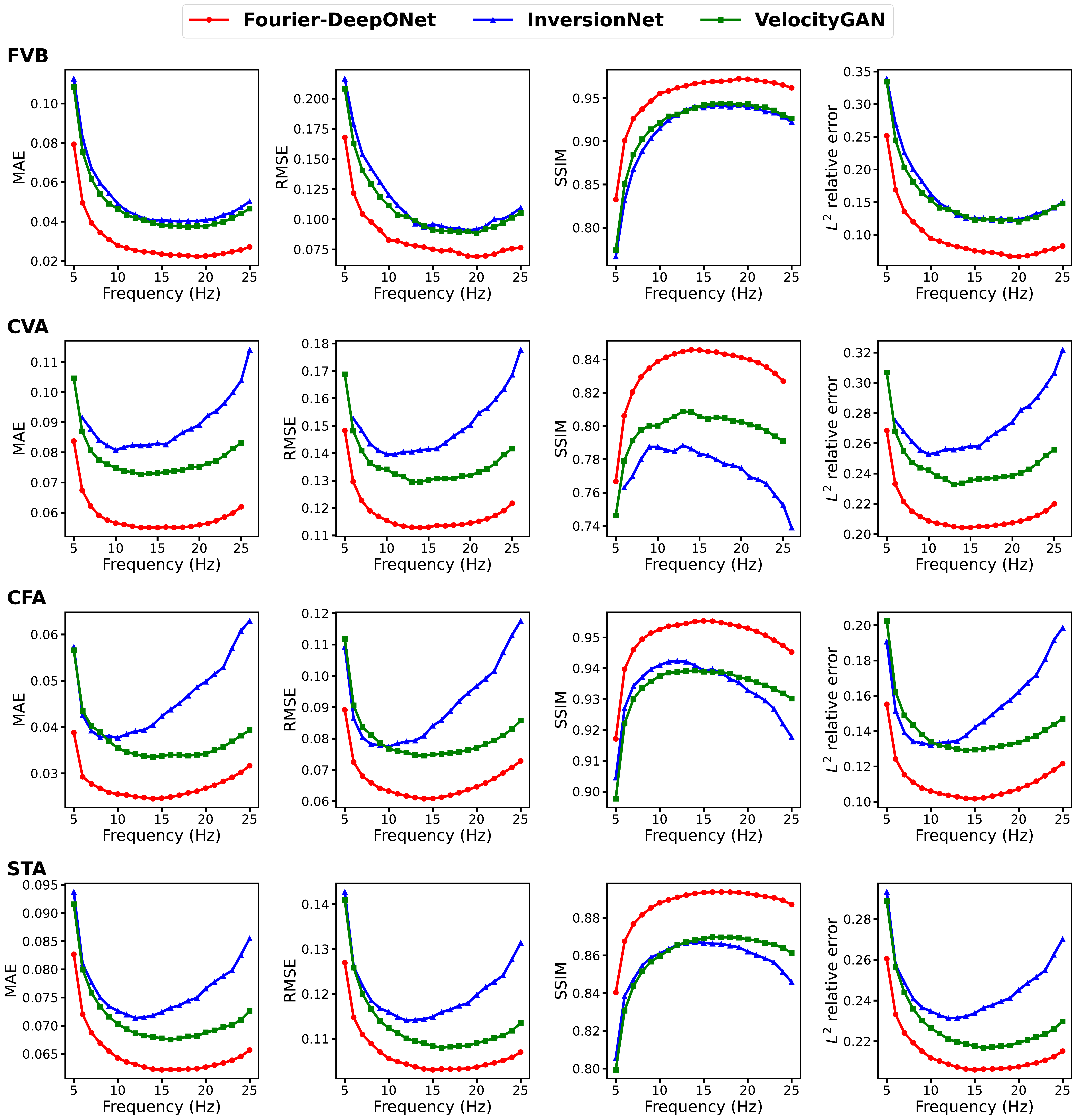}
    \caption{\textbf{Performance of three methods on four datasets (FVB, CVA, CFA, and STA) of different sources frequencies.} All three models are trained on FWI-F. Fourier-DeepONet outperforms the two baseline models on all datasets at any frequency ranging from 5 to 25 Hz.}
    \label{fig:frequency_add}
\end{figure}

Even compared with improved baseline models, Fourier-DeepONet still outperforms them across all datasets at any frequency between 5 and 25 Hz (Fig.~\ref{fig:frequency_add}). We note that compared with the results using OpenFWI, both InversionNet and VelocityGAN achieve better results at frequencies far from 15 Hz, but worse results at frequencies close to 15 Hz. This occurs because the source frequency of OpenFWI is fixed at 15 Hz, while the source frequency of FWI-F varies between 5 and 25 Hz. Generally, VelocityGAN has lower errors than InversionNet, indicating that it has better generalizability, but both of them are significantly worse than Fourier-DeepONet.

\subsection{Source locations}
\label{subsec:dev}

The source location also plays a critical role in FWI, as it directly affects the illumination of the subsurface and the quality of the resulting velocity model. The source location includes the position and distribution of the sources that generate the seismic waves used in the modeling process, which in turn impacts the accuracy, coverage, and convergence of the inversion. Source locations significantly influence the illumination of the subsurface structures. Different source positions result in different wave propagation paths, which can either improve or degrade the visibility of specific geological features. Proper source location selection ensures adequate illumination of the target area, leading to more accurate velocity models. In areas with complex geological structures, such as salt bodies, the source location becomes even more critical. Proper source locations can help address issues like wave multipathing, which can severely affect the inversion process's accuracy and convergence.

Existing FWI datasets, such as OpenFWI, only consider uniformly distributed sources. Although this might be a good choice in some ideal scenarios, it may not accurately represent the complexities and challenges associated with real-world FWI problems. In practical scenarios, the ideal uniform distribution of source locations may not be feasible due to topographical constraints, environmental concerns, or logistical limitations. Using datasets with only uniform source location distribution can limit the generalizability of the models trained on such datasets. Real-world FWI problems often require working with varying source locations to adequately illuminate and cover the subsurface structure.

\subsubsection{Overall performance}

The horizontal coordinates of the five sources in OpenFWI are 0, 172.5, 345, 517.5, and 690 m. To introduce variability in source locations, in datasets of FWI-L, we allow the sources to move within specified ranges, and the possible horizontal coordinates of the five sources are [0, 50], [122.5, 222.5], [295, 295], [467.5, 567.5], and [640, 690]~m. We train Fourier-DeepONet on FWI-L, while we use the baseline models (InversionNet and VelocityGAN) pretrained on OpenFWI from Ref.~\cite{dengopenfwi}. The performance of the three models on FWI-L is shown in Table~\ref{tab:loc}. In all of these four datasets, Fourier-DeepONet significantly outperforms the baseline models.

\begin{table}[htbp]
\centering
\caption{\textbf{Performance of three methods on FWI-L datasets.} Fourier-DeepONet significantly outperforms InversionNet and VelocityGAN in all datasets. Bold font indicates the best performance for each dataset.}
\label{tab:loc}
\begin{tabular}{cc|cccc}
\toprule
Dataset & Model & MAE & RMSE & SSIM & $L^2$ relative error\\    
\midrule
\multirow{3}{*}{FVB} & Fourier-DeepONet & \textbf{0.0211} & \textbf{0.0674} & \textbf{0.9709} & \textbf{0.0700} \\
& InversionNet & 0.2008 & 0.3460 & 0.6475 & 0.5816 \\
& VelocityGAN & 0.1995 & 0.3533 & 0.6328 & 0.5855 \\
\midrule
\multirow{3}{*}{CVA} & Fourier-DeepONet & \textbf{0.0531} & \textbf{0.1129} & \textbf{0.8468} & \textbf{0.2005} \\
& InversionNet & 0.1279 & 0.1854 & 0.7155 & 0.3310 \\
& VelocityGAN & 0.1082 & 0.1670 & 0.7423 & 0.3004 \\
\midrule
\multirow{3}{*}{CFA} & Fourier-DeepONet & \textbf{0.0232} & \textbf{0.0635} & \textbf{0.9571} & \textbf{0.1039} \\
& InversionNet & 0.1154 & 0.1828 & 0.8338 & 0.3490 \\
& VelocityGAN & 0.1215 & 0.1892 & 0.8310 & 0.3514 \\
\midrule
\multirow{3}{*}{STA} & Fourier-DeepONet & \textbf{0.0596} & \textbf{0.0999} & \textbf{0.8980} & \textbf{0.1994} \\
& InversionNet & 0.1842 & 0.2513 & 0.6834 & 0.5277 \\
& VelocityGAN & 0.1942 & 0.2651 & 0.6390 & 0.5546 \\
\bottomrule
\end{tabular}
\end{table}

\subsubsection{A particular test case for demonstration}

We further investigate the effect of source locations on prediction errors. As each source has its own location (i.e., in total five independent variables), it is not easy to visualize the results in a five-dimension space. To reduce the number of free variables, we fix the central source and move the other four sources equidistantly toward the center. In this setup, we only have one free variable: the shift distance from their default uniform locations for the four sources.

\begin{figure}[htbp]
    \centering
    \includegraphics[width=\textwidth]{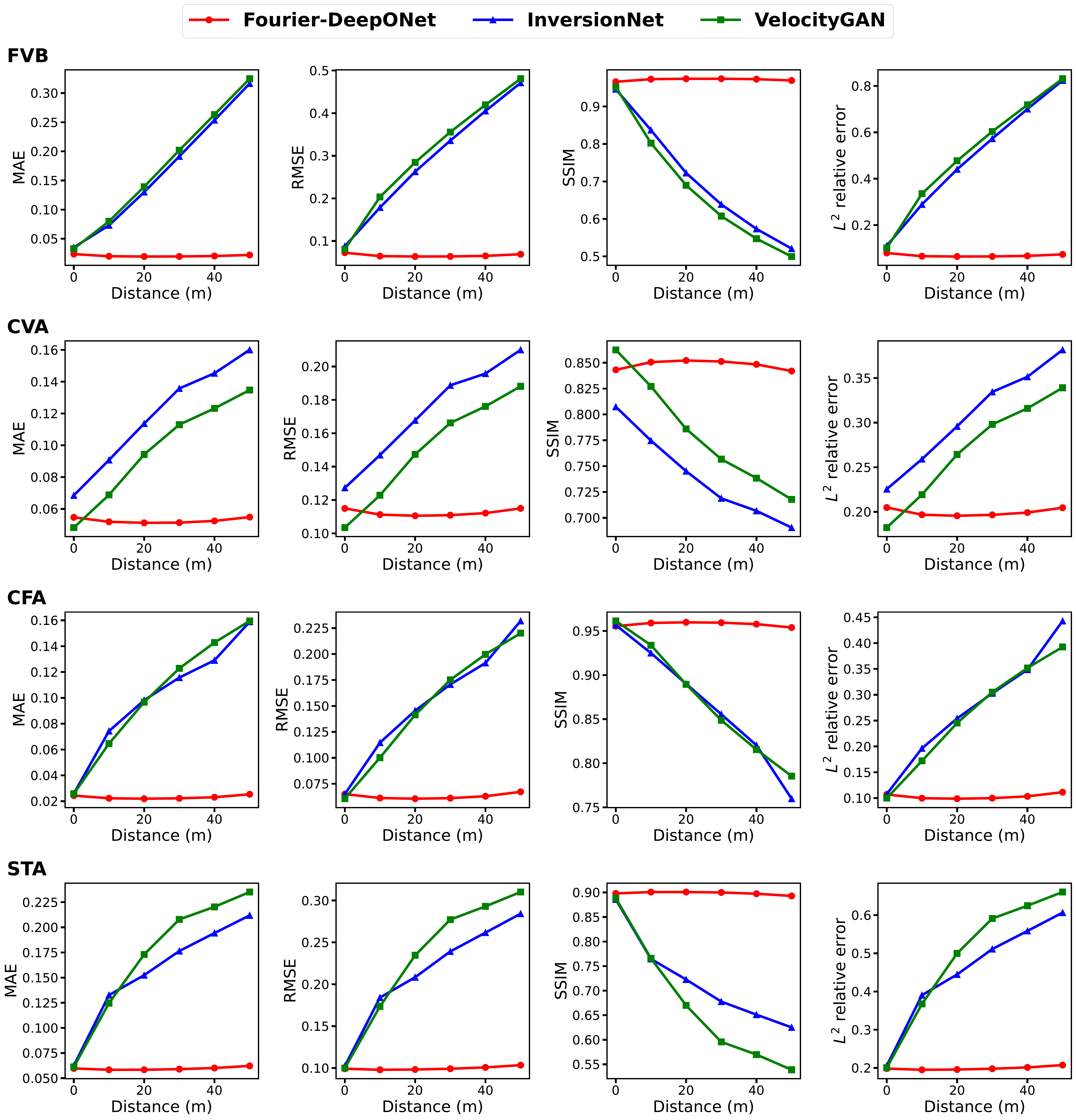}
    \caption{\textbf{Performance of three methods on four datasets (FVB, CVA, CFA, and STA) of different  distances from uniformly distributed sources.} Fourier-DeepONet is trained on FWI-F; InversionNet and VelocityGAN are trained on OpenFWI. Fourier-DeepONet performs well for all datasets when distance is from 0 to 50 m, while InversionNet and VelocityGAN can only give accurate predictions when distance is 0, i.e., the sources are uniformly distributed from left boundary to right boundary.}
    \label{fig:location}
\end{figure}

\begin{figure}[htbp]
    \centering
    \includegraphics[width=\textwidth]{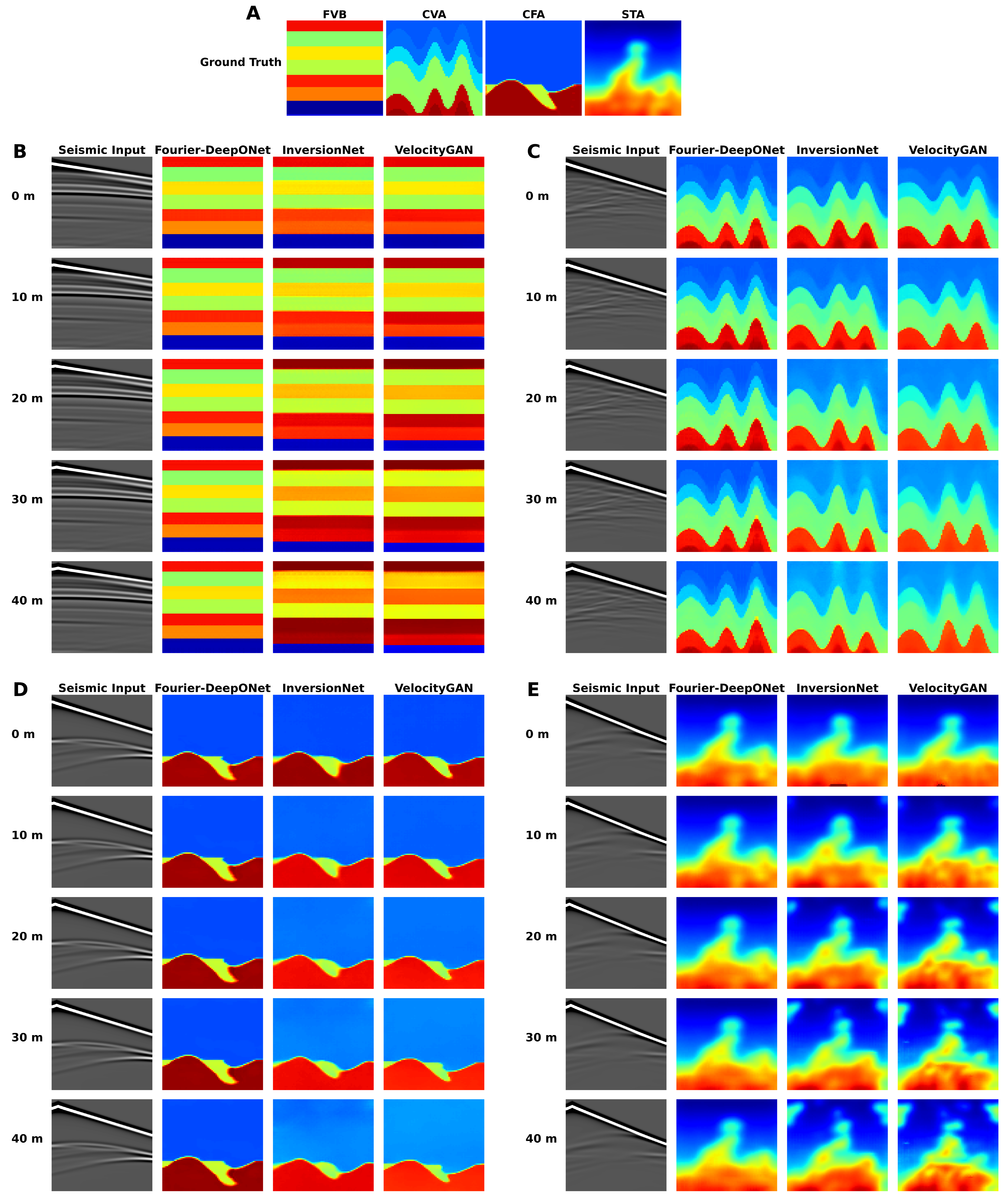}
    \caption{\textbf{Examples of velocity maps predicted by three methods on four datasets at different distances from uniformly distributed sources.} (\textbf{A}) Ground truth of examples from FVB, CVA, CFA, and STA datasets. (\textbf{B}) Predictions for the FVB case. (\textbf{C}) Predictions for the CVA case. (\textbf{D}) Predictions for the CFA case. (\textbf{E}) Predictions for the STA case. The seismic data comes from the source A in Fig.~\ref{fig:fwi}.}
    \label{fig:location_heatmap}
\end{figure}

Fig.~\ref{fig:location} shows the quantitative results of the three methods on FWI-L datasets with only one free variable. Fourier-DeepONet performs well for all datasets across a range of source locations, with low errors and high SSIM values, while InversionNet and VelocityGAN can only give accurate predictions when the shift distance is zero. Their prediction errors increase with the source location further deviating away. We also provide a visualization of the resulting prediction in Fig.~\ref{fig:location_heatmap}, which again demonstrates the strength of our proposed method by comparing it with InversionNet and VelocityGAN.

\subsection{Source frequencies and locations}
\label{subsec:f_loc}
In this section, we train and test Fourier-DeepONet on CVA dataset from FWI-FL.
Given the poor performance of the baseline models on both the FWI-F and FWI-L datasets, we refrain from testing them on the FWI-FL dataset.

\begin{figure}[htbp]
    \centering
    \includegraphics[width=\textwidth]{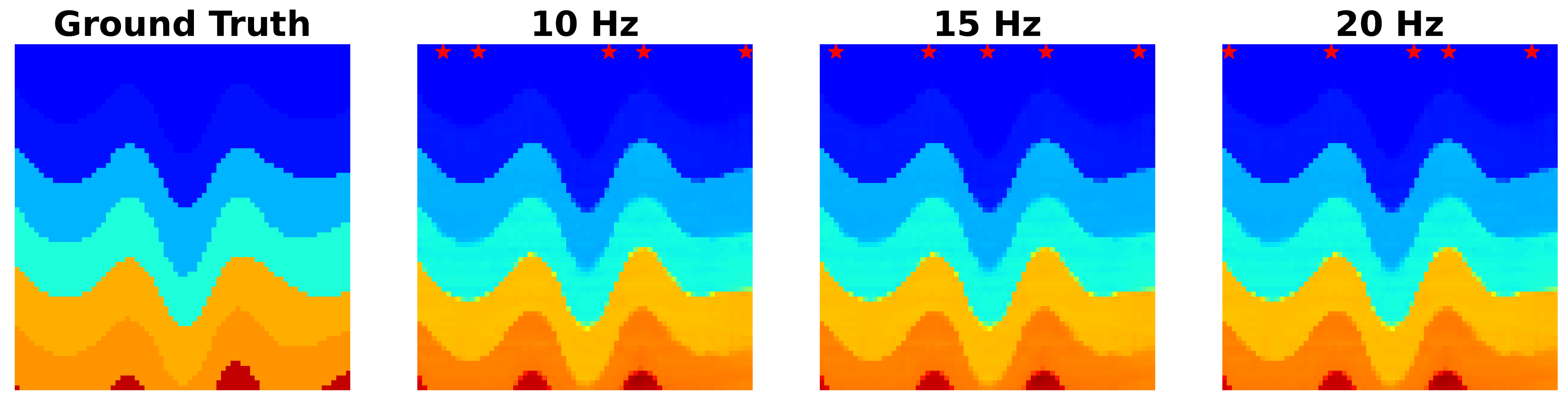}
    \caption{\textbf{Examples of velocity maps predicted by Fourier-DeepONet on CVA dataset from FWI-FL.} The source frequencies range from 10 to 20 Hz. The five red stars in the velocity map are the five point sources utilized to generate seismic data.}
    \label{fig:frequency_location}
\end{figure}

The MAE, RMSE, SSIM, and $L^2$ relative error of Fourier-DeepONet on testing dataset are 0.0663, 0.1276, 0.8176, and 0.2302, respectively. The accuracy is similar to that of Fourier-DeepONet on FWI-F or FWI-L datasets. Examples of velocity maps predicted by Fourier-DeepONet for source frequencies at 10, 15, and 20 Hz are shown in Fig.~\ref{fig:frequency_location}, where the five sources are randomly distributed on the surface. The Fourier-DeepONet provides satisfactory predictions with varying source frequencies and locations.

\subsection{Robustness evaluation}
\label{subsec:robustness}

We evaluate the robustness of the three models in three aspects: testing input data with Gaussian noise, testing input data with missing traces, and Ricker wavelet source with noise. We again leverage both velocity maps and seismic data from OpenFWI datasets, meaning uniformly distributed sources with fixed frequency at 15 Hz are used.

\subsubsection{Testing input data with noise}

Contaminating testing input data by adding noise is useful in evaluating the robustness of a model. Here, we use Gaussian noise to simulate noise that commonly occurs in real-world scenarios, such as sensor or measurement noise. The amount of added noise is controlled by specifying the standard deviation $\sigma$ of the Gaussian distribution. In our experiment, the range of $\sigma$ is from 0.01 to 1.

\begin{figure}[htbp]
    \centering
    \includegraphics[width=\textwidth]{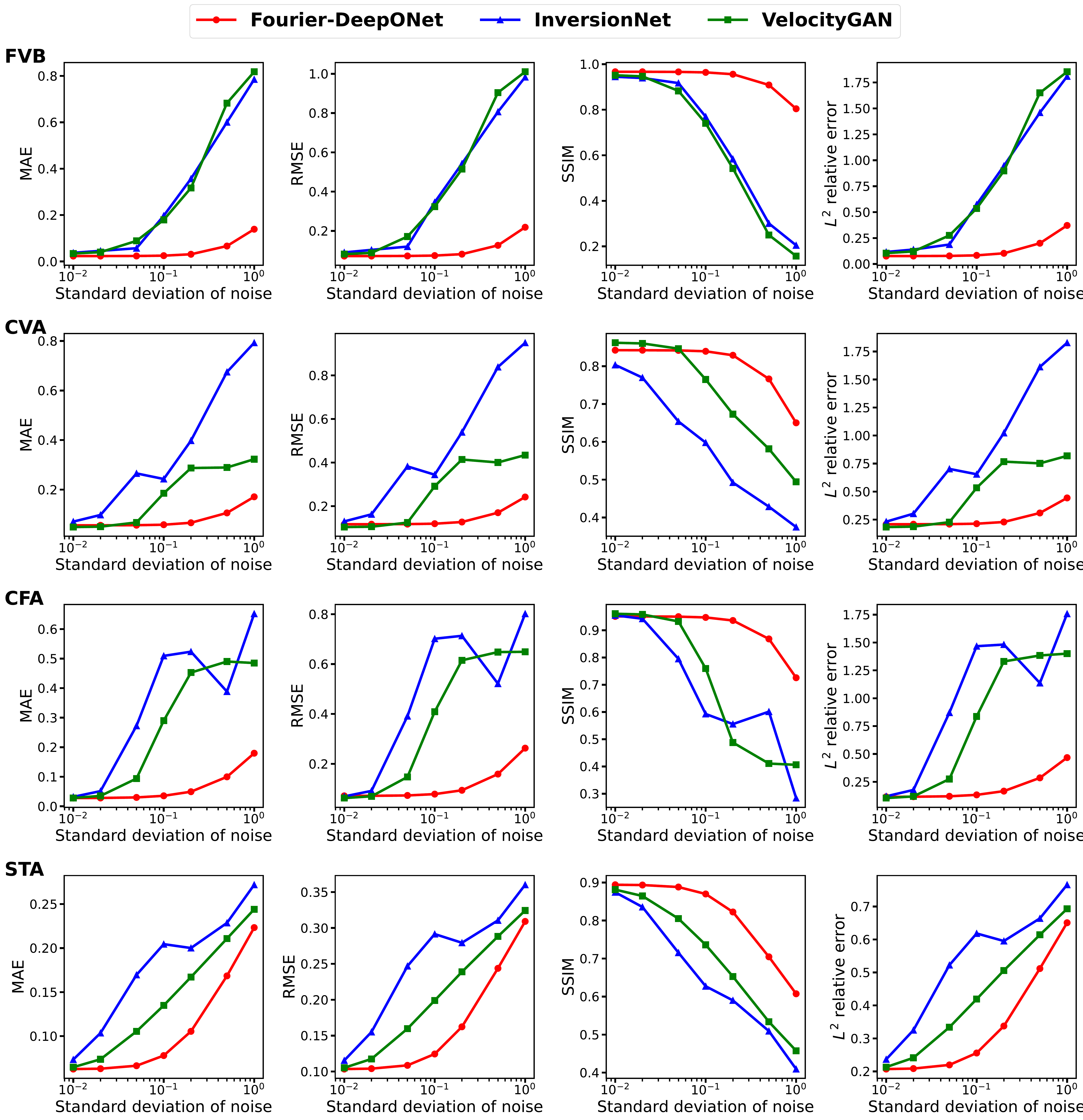}
    \caption{\textbf{Performance of three methods on four datasets (FVB, CVA, CFA, and STA) with different input noise levels during testing.} Gaussian noise of different standard deviations is added to the input seismic data during testing. Fourier-DeepONet shows significantly better robustness than the baseline models against noise.}
    \label{fig:robustness_noise}
\end{figure}

\begin{figure}[htbp]
    \centering
    \includegraphics[width=\textwidth]{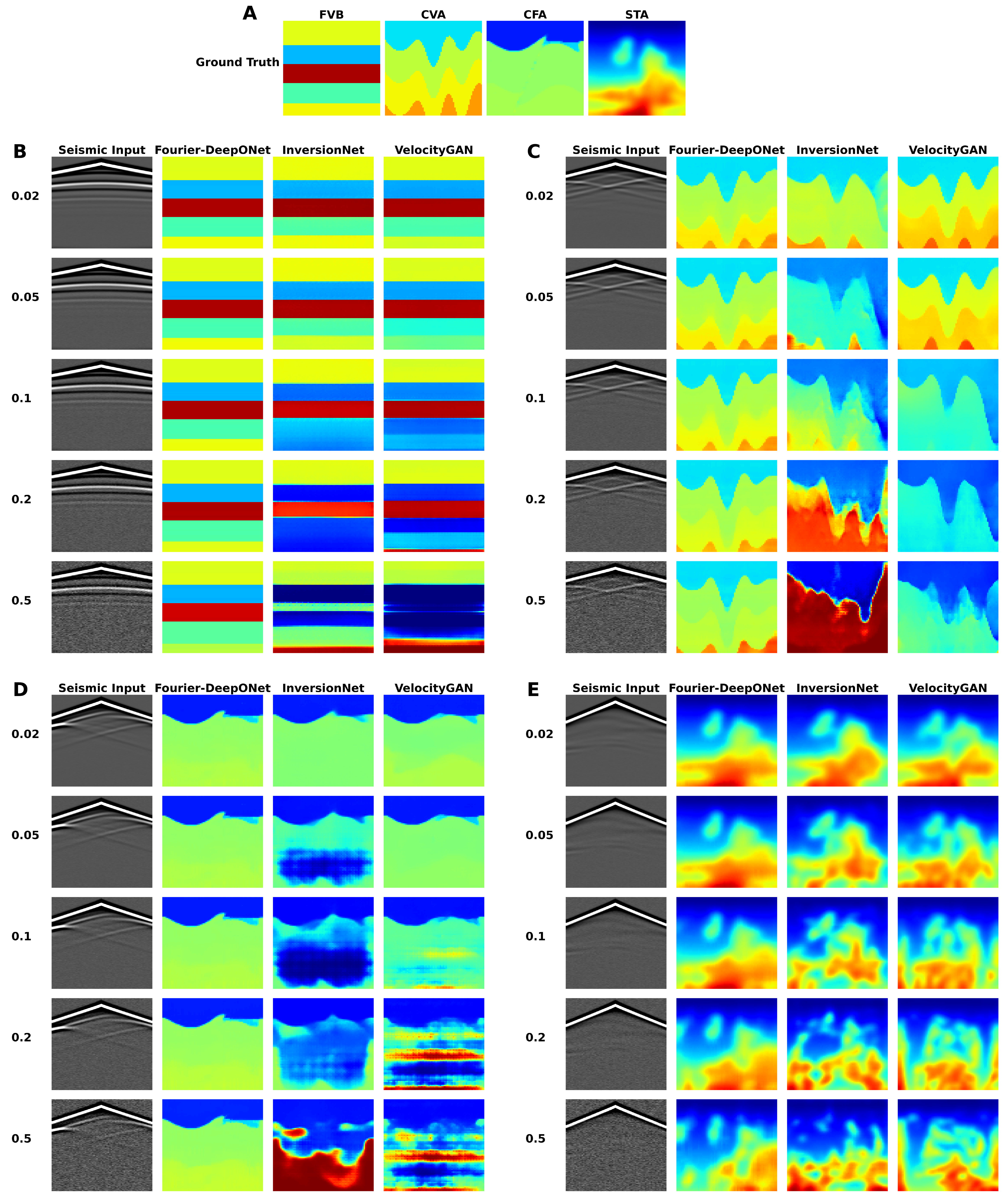}
    \caption{\textbf{Examples of velocity maps predicted by three methods on four datasets with various noisy seismic inputs.} (\textbf{A}) Ground truth of examples from FVB, CVA, CFA, and STA datasets. (\textbf{B}) Predictions for the FVB case. (\textbf{C}) Predictions for the CVA case. (\textbf{D}) Predictions for the CFA case. (\textbf{E}) Predictions for the STA case. The seismic data comes from the source C in Fig.~\ref{fig:fwi}. The numbers in Figure panels B--D represent the standard deviations of the Gaussian noise added to the seismic inputs.}
    \label{fig:noise_heatmap}
\end{figure}

We show that in all four datasets and different noise levels, Fourier-DeepONet performs much better than InversionNet and VelocityGAN (Fig.~\ref{fig:robustness_noise}). Even if the standard deviation $\sigma$ is up to 0.1, the accuracy of Fourier-DeepONet is almost unaffected, while InversionNet and VelocityGAN fail under this circumstance. We also show some prediction examples with various noisy seismic inputs on FVB, CVA, CFA, and STA datasets in Fig.~\ref{fig:noise_heatmap}.

\subsubsection{Testing input data with missing traces}

Missing traces can occur in FWI due to various reasons, such as irregularities in data acquisition, gaps in survey coverage, or loss of data during processing. While missing traces are common in real-world seismic datasets, the extent of their occurrence depends on the quality of the data acquisition and processing. The presence of missing traces can have a negative impact on the FWI results, potentially introducing artifacts or biases in the final subsurface property estimation. Therefore, the ability to deal with missing traces is crucial.

\begin{figure}[htbp]
    \centering
    \includegraphics[width=\textwidth]{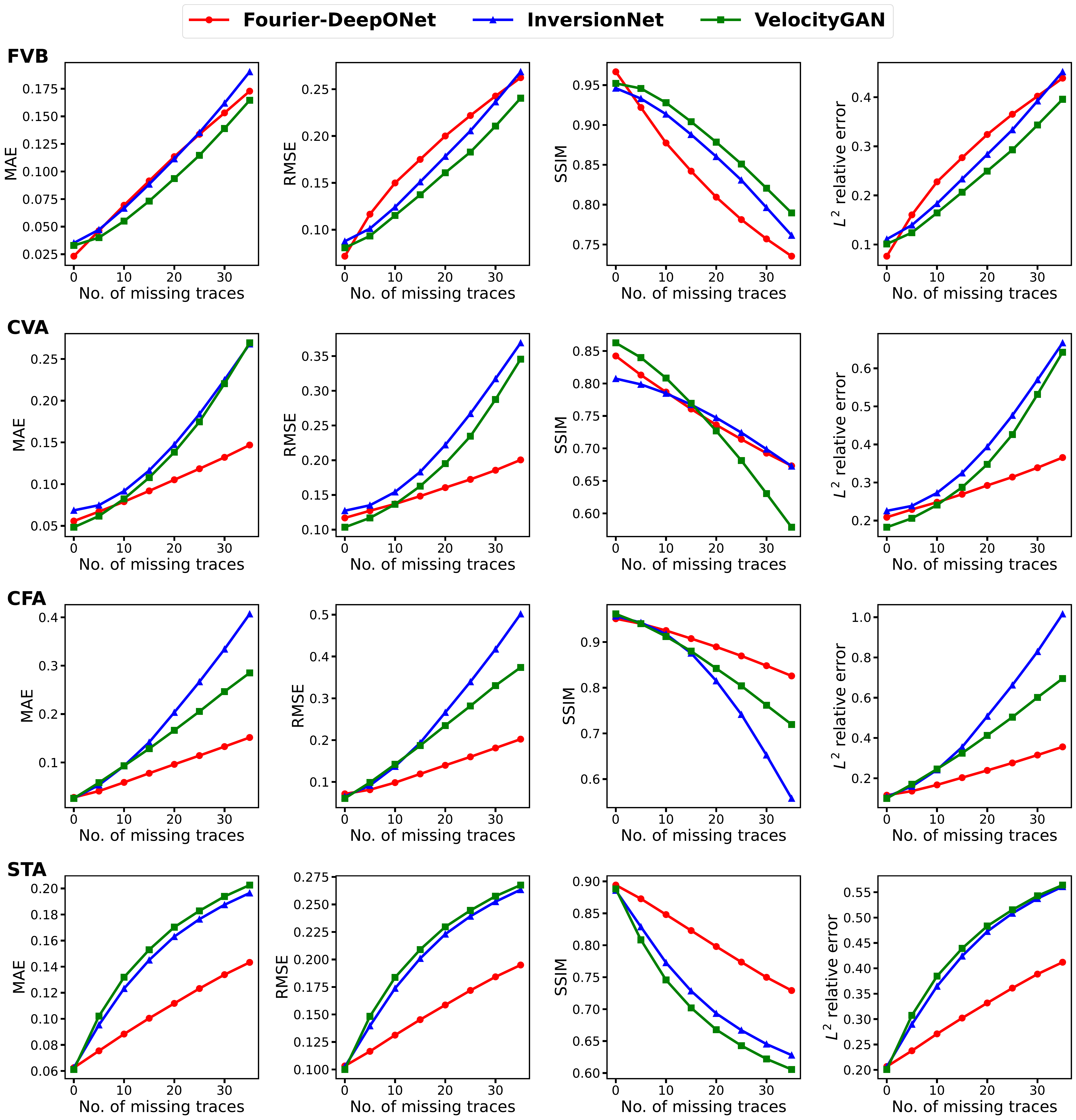}
    \caption{\textbf{Performance of three methods on four datasets (FVB, CVA, CFA, and STA) with different numbers of missing traces.} Missing traces of seismic data are filled with zero values during testing. Fourier-DeepONet shows better robustness than the baseline models against missing traces.}
    \label{fig:robustness_missing}
\end{figure}

\begin{figure}[htbp]
    \centering
    \includegraphics[width=\textwidth]{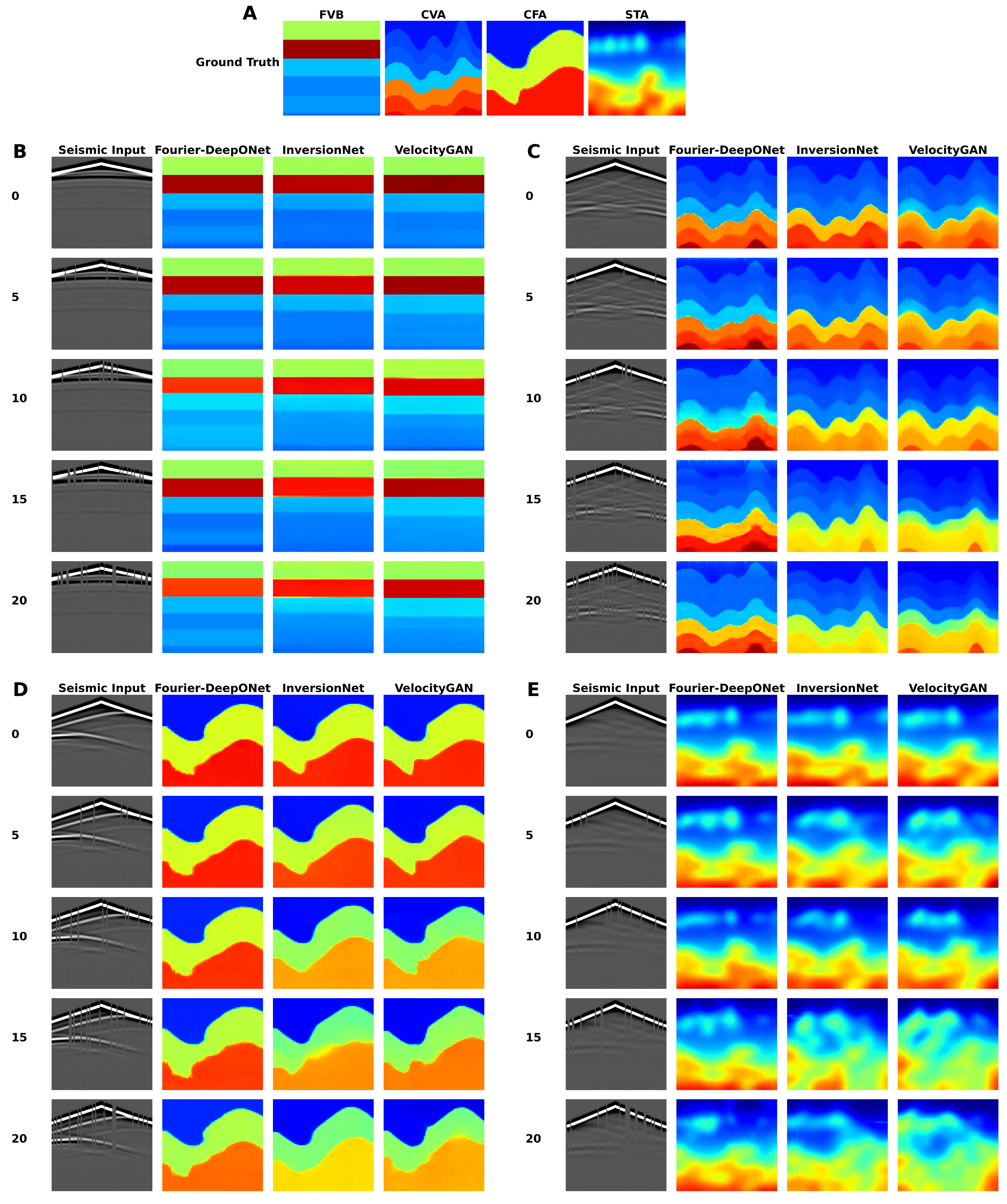}
    \caption{\textbf{Examples of velocity maps predicted by three methods on four datasets with different number of missing traces.} (\textbf{A}) Ground truth of examples from FVB, CVA, CFA, and STA datasets. (\textbf{B}) Predictions for the FVB case. (\textbf{C}) Predictions for the CVA case. (\textbf{D}) Predictions for the CFA case. (\textbf{E}) Predictions for the STA case. The seismic data comes from the source C in Fig.~\ref{fig:fwi}. The numbers in Figure panels B--D represent the numbers of missing traces. Missing traces of seismic data are filled with zero values during testing.}
    \label{fig:missing_heatmap}
\end{figure}

To evaluate the robustness of the three models in the face of missing traces, we randomly select some receivers to be treated as missing traces by setting the values on these missing traces to zero. The number of missing traces ranges from 5 to 35 out of a total of 70 receivers. In tests on CVA, CFA, and STA datasets, Fourier-DeepONet significantly outperforms InversionNet and VelocityGAN, especially when the number of missing traces increases (Fig.~\ref{fig:robustness_missing}). However, the three models have similar errors for FVB dataset, which may result from the simplicity of FVB dataset, and even when half of the receivers are missing, all three models can provide good predictions (SSIM $>$ 0.75). Examples of inversion results with various numbers of missing traces on FVB, CVA, CFA, and STA datasets are shown in Fig.~\ref{fig:missing_heatmap}.

\subsubsection{Ricker wavelet source with noise}
In our datasets, we use the Ricker wavelet as the amplitude of the source function. However, in practice, the sources that we impose on the surface are likely accompanied by noise. Here, we use Gaussian noise to simulate the noise added to the Ricker wavelet. The amount of added noise is controlled by specifying the standard deviation $\sigma$ of the Gaussian distribution. In our experiment, the range of $\sigma$ is from 0.001 to 0.1.

\begin{figure}[htbp]
    \centering
    \includegraphics[width=\textwidth]{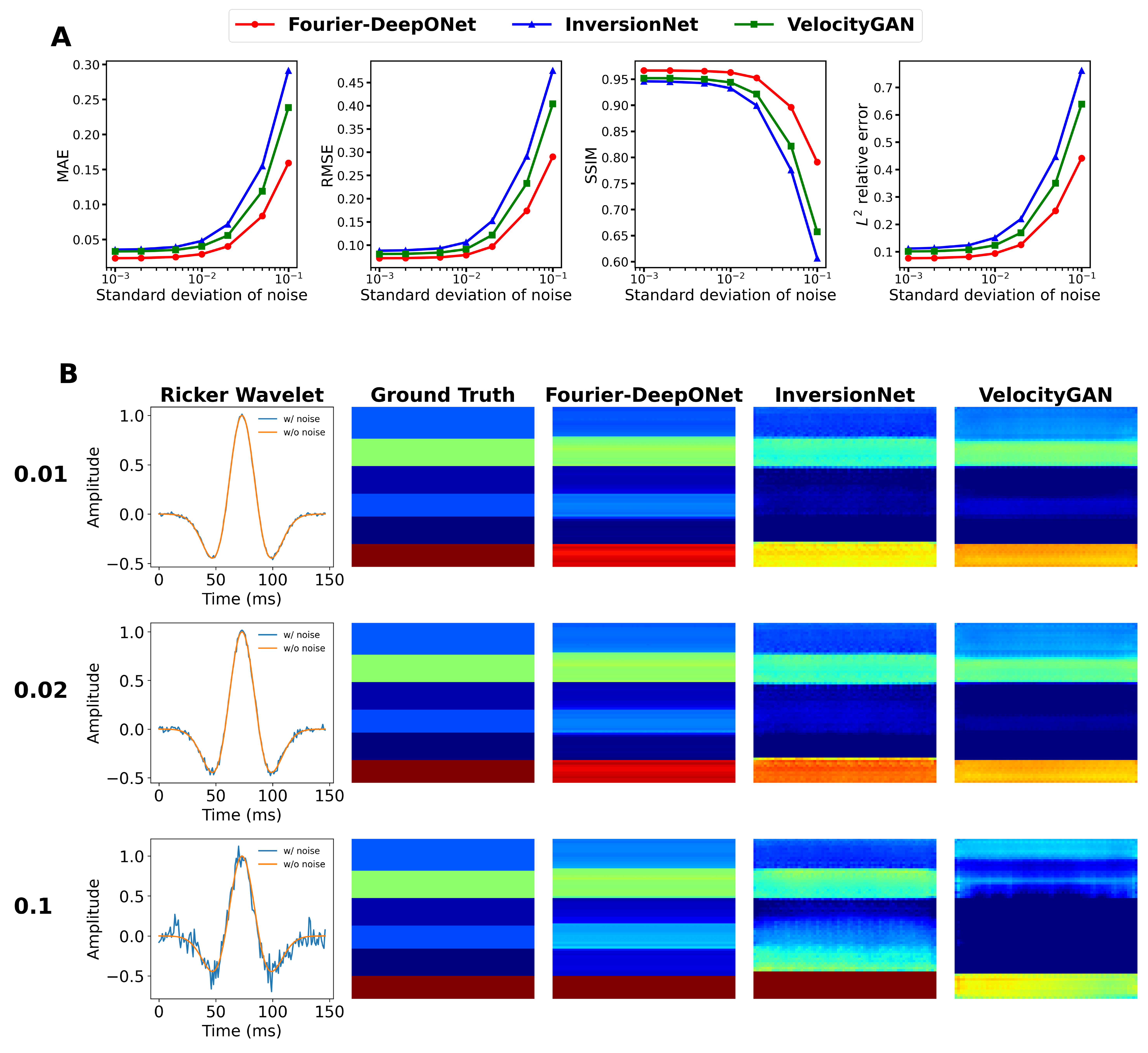}
    \caption{\textbf{Performance of three methods on FVB dataset  with different source noise levels during testing.} (\textbf{A}) Four metrics of three methods. Gaussian noise of different standard deviations is added to the sources during testing. Fourier-DeepONet shows better robustness than the baseline models against source noise. (\textbf{B}) Examples of velocity map predicted by three methods for noise level 0.01, 0.02, and 0.1.}
    \label{fig:source_noise}
\end{figure}

Our results show that at all noise levels, the Fourier-DeepONet consistently outperforms InversionNet and VelocityGAN (Fig.~\ref{fig:source_noise}). Even when the standard deviation $\sigma$ reaches 0.1, the Fourier-DeepONet can still provide satisfactory predictions, while both the InversionNet and VelocityGAN fail under these conditions.

\subsection{Discussion about the architecture of Fourier-DeepONet}
\label{subsec:architecture}

In this section, we provide more comparisons of our proposed Fourier-DeepONet and other networks.

\subsubsection{Comparison of different merger operations}
\label{subsec:operation}
In this study, we select pointwise multiplication after tensor broadcasting as the merger operation in Fourier-DeepONet. Here, we compare multiplication to the other two operations: addition and concatenation. For the concatenation operation, we allocate 48 channels to the branch net and 16 channels to the trunk net. This results in 64 channels after the branch and trunk nets are concatenated, which is the same as the number of channels for multiplication and addition. Comparisons of three merger operations on the FVB datasets from FWI-F and FWI-L are shown in Table~\ref{tab:operation}, respectively. In general, multiplication performs slightly better than the other two operations, but the differences are not significant. Following the convention of vanilla DeepONet, we choose multiplication as merger operation.

\begin{table}[htbp]
\centering
\caption{\textbf{Comparison of three merger operations on the FVB datasets from FWI-F and FWI-L.} Bold font indicates the best performance for different operations.}
\label{tab:operation}
\begin{tabular}{cc|cccc}
\toprule
Dataset & Operation & MAE & RMSE & SSIM & $L^2$ relative error\\   
\midrule
\multirow{3}{*}{FWI-F} & Multiplication & \textbf{0.0297} & \textbf{0.0849} & \textbf{0.9527} & \textbf{0.0960}\\
& Addition & 0.0359 & 0.0957 & 0.9393 & 0.1154 \\
& Concatenation & 0.0336 & 0.0868 & 0.9470 & 0.1079 
\\
\midrule
\multirow{3}{*}{FWI-L} & Multiplication & \textbf{0.0211} & 0.0674 & \textbf{0.9709} & \textbf{0.0700} \\
& Addition & 0.0218 & \textbf{0.0658} & 0.9708 & 0.0708 \\
&Concatenation & 0.0215 & 0.0690 & 0.9688 & 0.0726 \\
\bottomrule
\end{tabular}
\end{table}

\subsubsection{Comparison of Fourier and U-Fourier layers}
\label{subsec:fno_ufno}
In this study, we adopt one Fourier layer and three U-Fourier layers as the merger net. Here, we test the accuracy and efficiency of Fourier-DeepONets using only Fourier layers or U-Fourier layers (Table~\ref{tab:fno_ufno}). For the network with four Fourier layers, the testing error is the largest and training speed is the fastest. The network of four U-Fourier layers exhibits similar accuracy as the combination we adopted in this study. However, the network of four U-Fourier layers possesses more parameters and thus requires more training time. Therefore, the combination of one Fourier layer and three U-Fourier layers appears to be a better choice given the balance of performance and efficiency.

\begin{table}[htbp]
\centering
\caption{\textbf{Comparison of three Fourier-DeepONets on the CVA dataset from FWI-L.} The networks are trained on NVIDIA Tesla V100 SXM2 32 GB with a batch size of 32. Bold font indicates the best performance for different networks.}
\label{tab:fno_ufno}
\begin{tabular}{c|ccccc}
\toprule
 & MAE & RMSE & SSIM & $L^2$ relative error & Speed\\
  &&&&&(seconds/epoch)\\
\midrule
1 Fourier and 3 U-Fourier layers & 0.0531 & \textbf{0.1129} & \textbf{0.8468} & 0.2005 & 288\\
4 Fourier layers & 0.0648 & 0.1232& 0.8209 & 0.2182 & \textbf{218} \\
4 U-Fourier layers & \textbf{0.0506} & 0.1131 & 0.8429 & \textbf{0.1997} & 330\\
\bottomrule

\end{tabular}
\end{table}

\begin{figure}[htbp]
    \centering
    \includegraphics[width=\textwidth]{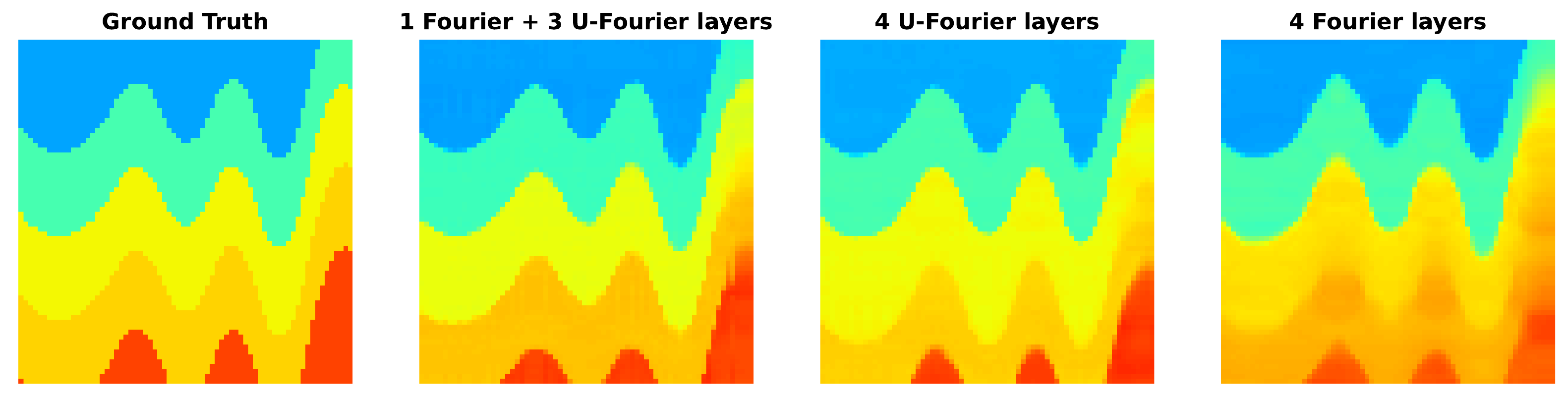}
    \caption{\textbf{An example of velocity map predicted by different Fourier-DeepONets on CVA dataset.}}
    \label{fig:fno_ufno}
\end{figure}

We show the prediction examples of the three networks in Fig.~\ref{fig:fno_ufno}. The network consisting of four Fourier layers performs the worst, exhibiting blurriness at the interfaces, due to the truncation of modes with higher frequencies. However, Fourier layers are faster than U-Fourier layers, as they do not contain the computationally expensive U-Net component. The U-Net within U-Fourier layers can capture the features of higher frequencies that are missed by Fourier layers, leading to both higher accuracy and increased computational costs.

\subsubsection{Comparison between Fourier-DeepONet and vanilla DeepONet}
\label{subsec:vanilla}
The vanilla DeepONet uses an inner product as the decoding mechanism to construct the output~\cite{lu2021learning}. Here, we compare the accuracy of Fourier-DeepONet and vanilla DeepONet. Both models have same training and testing datasets (FWI-L), number of parameters (28 million), training time (10 hours; NVIDIA Tesla V100 SXM2 32 GB). 

\begin{figure}[htbp]
    \centering
    \includegraphics[width=0.8\textwidth]{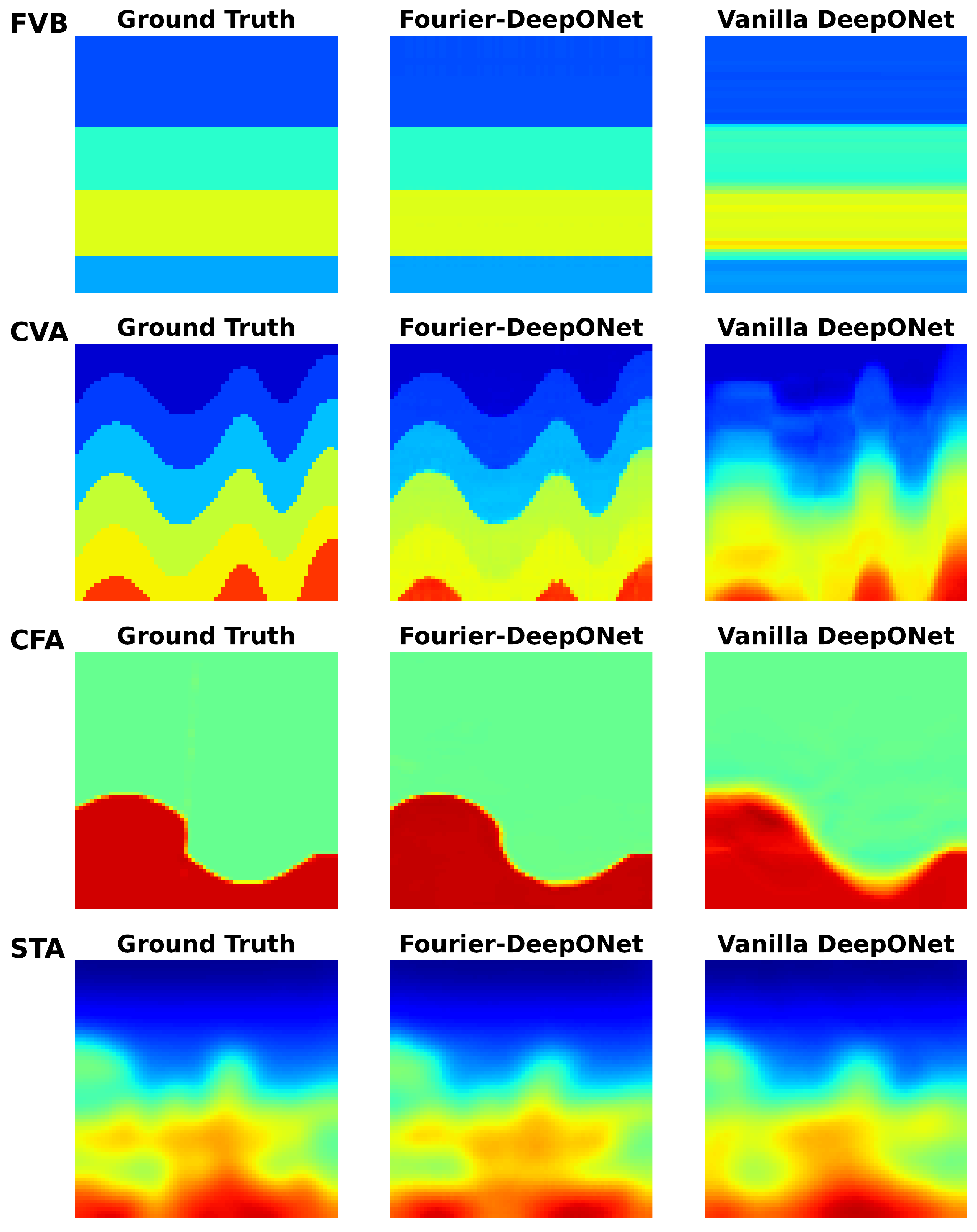}
    \caption{\textbf{Examples of velocity maps predicted by Fourier-DeepONet and vanilla DeepONet on the FVB, CVA, CFA, and STA datasets.} Vanilla DeepONet exhibit blurriness at the interfaces between adjacent layer.}
    \label{fig:vanilla_deeponet}
\end{figure}

The accuracy of the two models are shown in Table~\ref{tab:vanilla_deeponet}. Fourier-DeepONet outperforms vanilla DeepONet on all datasets, especially on FVB and CFA datasets. Prediction examples of the two models on four datasets are shown in Fig.~\ref{fig:vanilla_deeponet}. For FVB, CVA, and CFA datasets, the predictions from the vanilla DeepONet exhibit obvious blurriness at the interfaces between adjacent layers. On the other hand, vanilla DeepONet provides satisfactory visualizations on the STA dataset, probably because the velocity maps of STA dataset are smoother than those of the other datasets and do not contain sharp interfaces between adjacent layers.

\begin{table}[htbp]
\centering
\caption{\textbf{Comparison of Fourier-DeepONet and vanilla DeepONet on FWI-L datasets.} Fourier-DeepONet significantly outperforms vanilla DeepONet on all datasets. Bold font indicates the best performance for each dataset.}
\label{tab:vanilla_deeponet}
\begin{tabular}{cc|cccc}
\toprule
Dataset & Model & MAE & RMSE & SSIM & $L^2$ relative error\\    
\midrule
\multirow{2}{*}{FVB} & Fourier-DeepONet & \textbf{0.0211} & \textbf{0.0674} & \textbf{0.9709} & \textbf{0.0700} \\
& Vanilla DeepONet & 0.0558&0.1249&0.8657&0.1974 \\
\midrule
\multirow{2}{*}{CVA} & Fourier-DeepONet & \textbf{0.0531} & \textbf{0.1129} & \textbf{0.8468} & \textbf{0.2005} \\
& Vanilla DeepONet & 0.0715&0.1215&0.7817&0.2167  \\
\midrule
\multirow{2}{*}{CFA} & Fourier-DeepONet & \textbf{0.0232} & \textbf{0.0635} & \textbf{0.9571} & \textbf{0.1039} \\
& Vanilla DeepONet &0.0508&0.1131&0.8714&0.1942  \\
\midrule
\multirow{2}{*}{STA} & Fourier-DeepONet & \textbf{0.0596} & \textbf{0.0999} & \textbf{0.8980} & \textbf{0.1994} \\
& Vanilla DeepONet &0.0679&0.1044&0.8699&0.2120   \\

\bottomrule
\end{tabular}
\end{table}

\section{Conclusions}
\label{sec:conclusions}

In this paper, we developed the Fourier-enhanced deep operator network (Fourier-DeepONet) for solving FWI problems. Experiments show that Fourier-DeepONet is significantly more accurate than the baseline models (InversionNet and VelocityGAN) across a wide range of source parameters, indicating the superior generalizability of Fourier-DeepONet. The two baseline models can obtain relatively accurate results only when the sources are uniformly distributed and the frequencies are fixed at 15Hz. When the source frequencies or locations are not fixed, the $L^2$ relative errors of the two baseline models are about 10 times greater than that of Fourier-DeepONet. Additionally, we evaluate the robustness of all three models in data with Gaussian noise or missing traces and sources with Gaussian noise, and Fourier-DeepONet exhibits superior robustness compared to the other two models.



In the future, we will further examine cases where the sources are not Ricker wavelet. To this end, we will generate new seismic data with sources of various shapes, and then use the source amplitude values as trunk inputs. Moreover, Fourier-DeepONet could not only learn varying source locations but also adapt to different receiver locations. To accomplish this, we will take receiver locations as trunk inputs instead of source locations. On the other hand, to reduce the training dataset size, we could develop an unsupervised learning scheme by utilizing the governing equation of the problem~\cite{jin2021unsupervised, wang2021learning}.


\section*{Acknowledgement}

This work was supported by the U.S. Department of Energy (DOE) [DE-SC0022953], the Los Alamos National Laboratory (LANL) - Laboratory Directed Research and Development program under project number 20210542MFR, and the U.S. DOE Office of Fossil Energy's Carbon Storage Research Program via the Science-Informed Machine Learning to Accelerate Real Time Decision Making for Carbon Storage (SMART-CS) Initiative.

\bibliographystyle{unsrtnat}
\bibliography{main}

\end{document}